\documentclass{article}

\PassOptionsToPackage{numbers, compress}{natbib}
\usepackage[preprint]{neurips_2025}

\title{Cutting Smart, Reasoning Deep: Dynamic Difficulty-Aware CoT Distillation}

\title{Less is More: Efficient Reasoning Distillation through Difficulty-Aware Dynamic Prompting}

\title{Thinking Just Enough: Pruning Reasoning Chains via Difficulty-Aware Dynamic Prompting}

\title{Concise Reasoning, Big Gains: Pruning Long Reasoning Trace with Difficulty-Aware Prompting}

\author{
  {Yifan Wu}$^{1}$\quad
  {Jingze Shi}$^{2}$\quad
  {Bingheng Wu}$^{2}$\quad \\
  \textbf{Jiayi Zhang}$^{1,3}$\quad 
  \textbf{Xiaotian Lin}$^{1}$\quad
  \textbf{Nan Tang}$^{1}$\quad  
  \textbf{Yuyu Luo}$^{1}$ \\
  $^{1}$HKUST (Guangzhou) \quad
  $^{2}$Indepent Researcher\quad
  $^{3}$DeepWisdom
}
\newcommand{\eat}[1]{}
\usepackage[utf8]{inputenc} % allow utf-8 input
\usepackage[T1]{fontenc}    % use 8-bit T1 fonts
\usepackage{hyperref}       % hyperlinks
\usepackage{url}            % simple URL typesetting
\usepackage{booktabs}       % professional-quality tables
\usepackage{amsfonts}       % blackboard math symbols
\usepackage{nicefrac}       % compact symbols for 1/2, etc.
\usepackage{microtype}      % microtypography
\usepackage{xcolor}         % colors
\usepackage{enumitem}

\usepackage[utf8]{inputenc}
\usepackage[T1]{fontenc}
\DeclareUnicodeCharacter{03B8}{\ensuremath{\theta}}
\DeclareUnicodeCharacter{03C0}{\ensuremath{\pi}}
\DeclareUnicodeCharacter{2208}{\ensuremath{\in}}
\DeclareUnicodeCharacter{2202}{\ensuremath{\partial}}
\DeclareUnicodeCharacter{2207}{\ensuremath{\nabla}}
\DeclareUnicodeCharacter{03BB}{\ensuremath{\lambda}}
\DeclareUnicodeCharacter{221A}{\ensuremath{\surd}}   % 或者只映射成 \surd
\DeclareUnicodeCharacter{2248}{\ensuremath{\approx}}
\DeclareUnicodeCharacter{2264}{\ensuremath{\leq}}
\DeclareUnicodeCharacter{0394}{\ensuremath{\Delta}}
\usepackage[color,matrix,arrow,all]{xy}
\usepackage{pifont}
\usepackage{amsmath}
\usepackage{tcolorbox}
\tcbuselibrary{breakable}
\usepackage{url}
\usepackage{wrapfig}
\usepackage{caption}
\usepackage{listings}
\usepackage{makecell}
\usepackage{xparse}
\usepackage{wrapfig}
\usepackage{tcolorbox}

\usepackage{epsfig}
\usepackage{multirow}
\usepackage{bbm}

\definecolor{shadecolor}{RGB}{220,220,220}

\definecolor{inputcolor}{RGB}{255,139,35}
\definecolor{outputcolor}{RGB}{120,212,252}
\definecolor{embedcolor}{RGB}{254,127,156}
\definecolor{maskcolor}{RGB}{122,128,255}
\definecolor{ecolor}{RGB}{58,149,54}

\definecolor{highcolor}{RGB}{255,153,153}
\definecolor{midcolor}{RGB}{255,204,204}
\definecolor{lowcolor}{RGB}{204,229,255}

\usepackage{tikz}
\usetikzlibrary{shapes,snakes}
\usetikzlibrary{calc}

\usepackage{algorithm} 
\usepackage{algorithmic}  
\usepackage[algo2e]{algorithm2e} 

\usepackage[export]{adjustbox}% http://ctan.org/pkg/adjustbox

\definecolor{green}{RGB}{0,128,0}

\definecolor{yellow}{RGB}{255,200,18}

\newcommand{\stab}{\vspace{1.2ex}\noindent}

\newcommand{\bi}{\begin{itemize}}
\newcommand{\ei}{\end{itemize}}

\newcommand{\be}{\begin{enumerate}}
\newcommand{\ee}{\end{enumerate}}
\newcommand{\beqn}{\begin{eqnarray*}}
\newcommand{\eeqn}{\end{eqnarray*}}

\newcommand{\stitle}[1]{\stab\noindent{\bf #1}}

\newcommand{\ie}{\textit{i.e.,}\xspace}
\newcommand{\eg}{\textit{e.g.,}\xspace}

\newcommand{\model}{{Liter}\xspace}

\newcommand{\data}{{LiteCoT}\xspace}

\usepackage{tcolorbox}
\usepackage{colortbl}
\usepackage{soul}
\definecolor{c1}{cmyk}{0,0.6175,0.8848,0.1490}
\definecolor{c2}{cmyk}{0.1127,0.6690,0,0.4431}
\definecolor{c3}{cmyk}{0.3081,0,0.7209,0.3255}
\definecolor{c4}{cmyk}{0.6765,0.2017,0,0.0667}
\definecolor{c5}{cmyk}{0,0.8765,0.7099,0.3647}

\newtcbox{\hlprimarytab}{on line, rounded corners, box align=base, colback=c3!10,colframe=white,size=fbox,arc=3pt, before upper=\strut, top=-2pt, bottom=-4pt, left=-2pt, right=-2pt, boxrule=0pt}
\newtcbox{\hlsecondarytab}{on line, box align=base, colback=red!10,colframe=white,size=fbox,arc=3pt, before upper=\strut, top=-2pt, bottom=-4pt, left=-2pt, right=-2pt, boxrule=0pt}
\newtcolorbox[]{finding}[0]{colback=gray!10, colframe=black, width=\columnwidth, boxrule=0.4pt, 
left=0mm, right=0mm, top=0mm, bottom=0mm, before skip=3pt, after skip=3pt, sharp corners}

\makeatletter
    \newcommand\figcaption{\def\@captype{figure}\caption}
    \newcommand\tabcaption{\def\@captype{table}\caption}
\makeatother

\tikzstyle{mybox} = [draw=black, fill=black!5, thick,
   rectangle, rounded corners, inner sep=0pt, inner ysep=6pt]
\tikzstyle{fancytitle} =[fill=black, text=white]

\NewDocumentCommand{\nan}{ mO{} }{\textcolor{blue}{\textsuperscript{\textit{Nan}}\textsf{\textbf{\small[#1]}}}}

\NewDocumentCommand{\yuyu}{ mO{} }{\textcolor{green}{\textsuperscript{\textit{Yuyu}}\textsf{\textbf{\small[#1]}}}}

\NewDocumentCommand{\yifan}{ mO{} }{\textcolor{brown}{\textsuperscript{\textit{Yifan}}\textsf{\textbf{\small[#1]}}}}
\NewDocumentCommand{\lutao}{mO{}  }{\textcolor{orange}{\textsuperscript{\textit{lutao}}\textsf{\textbf{\small[#1]}}}}

\usepackage{marginnote}
\let\oldmarginpar\marginpar
\renewcommand\marginpar[1]{\-\oldmarginpar[\raggedleft\footnotesize #1]%
	{\raggedright\footnotesize\color{blue} #1}} % 注释文字用红色footnote 大小
\usepackage{marginnote}
\let\oldmarginnote\marginnote
\renewcommand\marginnote[1]{\-\oldmarginnote[\raggedleft\footnotesize #1]%
	{\raggedright\footnotesize\color{blue} #1}} %
\begin{document}

\maketitle

\begin{abstract}
Existing chain-of-thought (CoT) distillation methods can effectively transfer reasoning abilities to base models but suffer from two major limitations: excessive verbosity of reasoning traces and inadequate adaptability to problem difficulty. Long reasoning traces significantly increase inference costs, and uniform-length solutions prevent base models from learning adaptive reasoning strategies.
To address these issues, we propose a difficulty-aware prompting (\textbf{DAP}) method to dynamically shorten reasoning traces without performance loss.
In our approach, a large teacher model first judges each problem's difficulty and then rewrites its reasoning traces to an appropriate shorter length, yielding concise yet complete reasoning traces.
Leveraging the DAP pipeline, we curate a distilled dataset called \textbf{LiteCoT} consisting of 100K concise reasoning examples, with solutions averaging only 720 tokens (an order of magnitude shorter than typical CoTs).
Using LiteCoT, we distilled a new family of reasoning models called \textbf{Liter} (1.5B, 7B, and 32B) based on the Qwen2.5 architecture. Experiments show that a student model fine-tuned on just 100K of these difficulty-pruned CoT samples outperforms a model distilled on 800K original Long CoT samples, while significantly reducing training and inference costs. Our method also generalizes well: across 11 diverse benchmarks, the shorter difficulty-aware CoTs achieve equal or better accuracy than Long chains, using far fewer tokens. For example, on the challenging AIME24 exam, our approach reaches 74.2\% Pass@1 using only about 5K inference tokens, surpassing other methods that consume many more tokens. Our code and data are available at \url{https://github.com/Evanwu1125/LiteCoT}. 
% Experiments show that a model trained on only 100K difficulty-pruned CoT samples outperforms a model distilled on 800K original long CoT samples, while significantly reducing training and inference costs. Our method also demonstrates strong generalization: across diverse benchmarks, the shorter difficulty-aware CoT yields equal or better accuracy than long chains, all with far fewer tokens. 
% On the AIME24 benchmark, our method achieves 74.2\% Pass@1 accuracy using only about 5K inference tokens, which is significantly better than the performance achieved by other comparative methods when consuming much more tokens. 

% \yuyu{@Yifan, please double-check expr results}.
% \yuyu{Our code and data are available at  ???}

% This effectively solves the main limitations of the existing distillation paradigm in training efficiency and resource consumption. The main contribution of this paper is to systematically analyze the cost bottleneck of direct distillation, propose a novel adaptive difficulty reasoning distillation framework to optimize training resources, design a matching prompt template, and verify its significant effectiveness in reducing training costs, improving efficiency and model performance through experiments.
\end{abstract}

\section{Introduction}
\label{sec:intro}

Reasoning language models (RLMs) like DeepSeek-R1~\citep{deepseekr1} and OpenAI-o1~\citep{GPT-o1} have achieved impressive results on complex tasks (\eg math reasoning) by generating explicit chain-of-thought (CoT) traces.
Fine-tuning smaller base models on such CoT traces, \ie a form of model distillation, can effectively transfer reasoning capabilities.
For example, using the \textit{input–output} traces from a powerful reasoner as supervised training data can substantially boost a smaller model's performance~\citep{ye2025limo,llmsarereasoningteachers,liu2025advances}.

% Recently, reasoning language models (RLMs) represented by DeepSeek R1 and GPT-o1 have achieved impressive results on complex tasks like mathematical reasoning problems by producing explicit chain-of-thought (CoT) reasoning traces. 
% Fine-tuning base models on these CoT traces, which represent a form of \textit{model distillation}, has proven effective for transferring reasoning capabilities~\citep{deepseekr1, efficientKS}.
% For example, simply using input–output pairs from an advanced reasoner as supervised fine-tuning data can substantially boost a smaller model's reasoning performance~\citep{ye2025limo,llmsarereasoningteachers}.

% Reasoning language models have demonstrated impressive reasoning capabilities through reinforcement learning from verifiable data. Models like DeepSeek R1 \cite{deepseekr1} and GPT-o1 \cite{GPT-o1} achieve remarkable performance on complex reasoning tasks by producing explicit chain-of-thought (CoT) traces. Notably, these reasoning traces can be used to enhance other models' capabilities: fine-tuning base models on these CoT outputs effectively transfers reasoning abilities \cite{deepseekr1, add}, demonstrating the potential of model distillation in propagating advanced reasoning skills.

\begin{figure}[t!]
    \centering
    \vspace{-1em}
    \includegraphics[width=\textwidth]{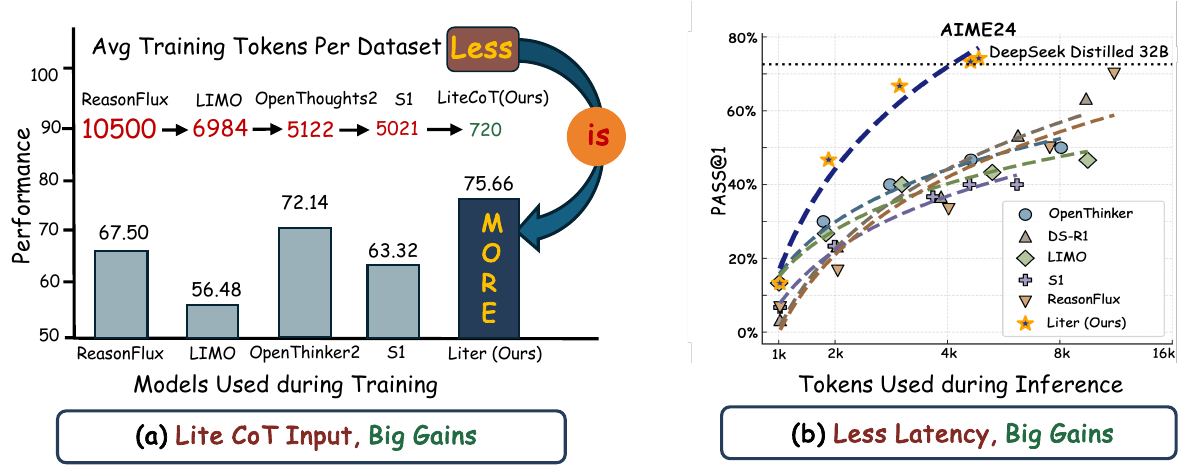}
    \vspace{-1em}
    \caption{(a) Training tokens comparison of LiteCoT with prior distillation methods (32B model); (b) Inference efficiency and accuracy on the challenging AIME24 benchmark (32B model).
    Our \model models trained with LiteCoT achieve higher accuracy (74.2\% Pass@1) at substantially reduced inference token counts compared to prior methods.}
    \label{fig:intro_figure}
     \vspace{-2em}
\end{figure}

% \lun{Better says, Limitations on Existing Methods/Strategies, I think you are discussing diff. strategies on model distillation}

\textbf{Existing CoT Distillation Strategies.} Researchers have explored various strategies to improve CoT distillation data quality. 
For example, S1~\citep{s1} filtered the 59K initial data collected from multiple angles, such as difficulty and diversity, and finally obtained 1K high-quality reasoning traces for distillation. 
Light R1~\citep{lightr1} filtered 100M raw \textit{question–answer} pairs down to 70K high-quality reasoning traces for distillation, and applied a two-stage model fine-tuning process.
ReasonFlux~\citep{yang2025reasonflux} designed templates to synthesize additional long reasoning examples from existing data.
LIMO~\citep{ye2025limo} used a classifier LLM to select only the most challenging questions for distillation data.
These efforts have improved data quality in some aspects, but two key \textbf{limitations} remain in practice. 
First, even after careful filtering or synthesis, the CoT outputs from the large teacher (\ie a strong LLM) are very long (\eg the maximum can contain 32K tokens), which will incur heavy computational cost and time to train on. 
Second, current reasoning traces distilled methods lack adaptability in reasoning trace length. In practice, simple problems might be solvable with only a few intuitive steps, whereas truly complex questions demand more elaborate multi-step reasoning. However, existing strategies do not differentiate solution lengths by problem difficulty.
Large models often ``overthink'' even simple queries~\citep{2+3}, and if all distilled solutions are uniformly verbose, the student model never learns to adjust reasoning detail to the task at hand~\citep{stopoverthinking, chen2025seal, overreasoning}. In short, existing CoT distillation methods tend to produce \textit{one-size-fits-all} long traces, failing to adapt to actual problem complexity.

Therefore, \textbf{our goal} is to construct a new distilled reasoning dataset that addresses these challenges by being both concise and difficulty-adaptive.
Shorter, more concise reasoning traces can cut down training time and token usage, while difficulty-adaptive traces mean the model sees only as much reasoning as necessary, \ie brief for easy tasks and more thorough for harder ones. Intuitively, models trained on such tailored data can reason more efficiently by avoiding redundant computations on simple tasks and effectively allocating resources to complex ones.

\begin{figure}[t!]
    \centering
    \includegraphics[width=\textwidth]{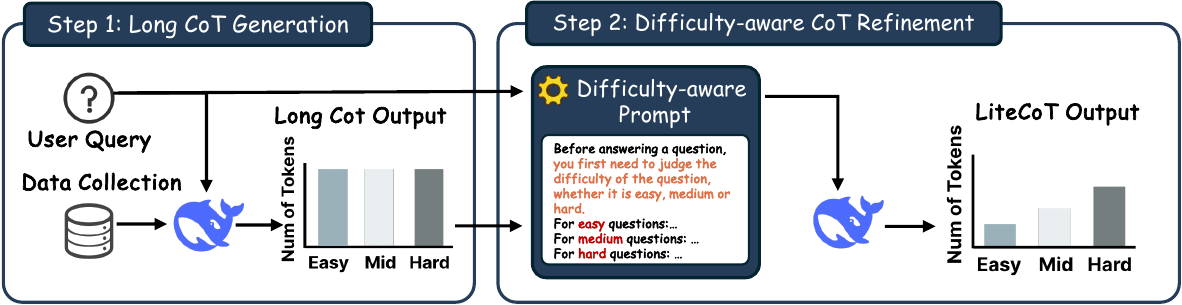}
    \vspace{-1.5em}
    \caption{Overview of the DAP pipeline. First, the DeepSeek R1 model generates initial Long CoT data from collected user queries. Then, a difficulty-aware prompt guides the model to rewrite concise CoT outputs, adapting reasoning length dynamically based on problem difficulty.}
    \label{fig:method_figure}
\end{figure}

\textbf{Simple Yet Effective CoT Distillation Method: Difficulty-Aware Prompting (DAP).}  
To achieve this, our key idea is to use a powerful teacher model's ability to both control reasoning length and assess problem difficulty when properly prompted. 
As shown in Figure~\ref{fig:method_figure}, we first design prompt templates for three levels of problem difficulty (\textit{simple}, \textit{medium}, and \textit{complex}), which have different requirements on the structure and level of detail of the output reasoning path. We then take the long reasoning trace in the original distilled data as input, and require DeepSeek-R1 to rewrite the lengthy original reasoning chain into a new reasoning path that is simpler and has a more optimized length, combining the structured instructions in the prompt corresponding to the selected difficulty. 
% Please refer to Section~\ref{DAP Pipeline} for details.

\textbf{LiteCoT Dataset.}
Leveraging the proposed DAP pipeline, we constructed a new distilled dataset called \textbf{\data}, containing 100K concise, difficulty-adaptive reasoning samples.
Each solution in \data averages approximately \textit{720} tokens, representing an order-of-magnitude reduction in CoT length compared to existing distillation datasets, \eg S1~\cite{s1} and OpenThoughts~\cite{open_thoughts}, which typically contain between \textit{5K} and \textit{10K} tokens per sample.
As shown in Figure~\ref{fig:intro_figure}(a), LiteCoT significantly reduces training token counts while achieving the best results, clearly demonstrating the effectiveness of concise, structured reasoning inputs.

\textbf{Liter Models and Experiments.}
Using \data, we distilled a new family of reasoning models named \textbf{\model} (1.5B, 7B, 14B and 32B) based on Qwen2.5 architecture.
Comprehensive evaluations across 11 diverse benchmarks demonstrate the superiority of models trained on our compact LiteCoT dataset. Remarkably, models distilled on the LiteCoT dataset (100K examples) consistently outperform those distilled on 800K original Long CoT examples, despite the latter having eight times more training data and substantially greater token exposure. Specifically, 
as shown in Figure~\ref{fig:intro_figure}(b), our \model achieves a Pass@1 accuracy of 74.2\% using only around 5K inference tokens per example on the challenging AIME24 benchmark. This performance significantly surpasses previous methods, which require substantially more tokens to achieve comparable or even lower accuracy.
\section{Related Work}
% \yuyu{full related work, move to appendix, we keep concise one in the paper.}

% \subsection{Knowledge Distillation in Large Language Models}
% Knowledge distillation was originally designed to make the student model fit the probability distribution of the teacher model~\citep{}, while in the era of large language models~\citep{}, a more common approach is to let the student model directly imitate the complete output sequence of the teacher model~\cite{}. This supervised fine-tuning(SFT) distillation paradigm based on sequence imitation is effective because strong reasoning models such as DeepSeek R1 can generate high-quality chain-of-thought(CoT) containing rich reasoning paths~\cite{}. Related studies have also shown that fine-tuning based on these detailed CoT data can significantly improve the performance of small models on tasks such as mathematical reasoning~\cite{}. What's more, Ye et al.~\cite{} analysed the quality of distilled data and proved that small number of high quality reasoning data can significantly improve base model's reasoning ability. Although this distillation method has been proven to be an effective training paradigm, the original CoT output from the reasoning model often contains a lot of redundant information, which directly leads to the expansion of training data and significantly increased training costs.

% \yuyu{@co-authors: 1. please fix the cite issues; 2. please add paragraph title}

\textbf{LLM Distillation.} Knowledge distillation in large language models has evolved from traditional probability distribution matching to sequence-level imitation, where student models directly learn from teacher outputs. Models like DeepSeek-R1~\citep{deepseekr1}, OpenAI-o1~\citep{GPT-o1} generate high-quality chain-of-thought (CoT) reasoning paths that, when used as training data, significantly improve smaller models' reasoning abilities~\citep{open_thoughts, sky_t1_2025}. Recent work by~\citep{ye2025limo, yang2025reasonflux,guan2025rstar,zhang2024aflow, xiang2025self, lin2025lead} demonstrated that a small collection of high-quality reasoning examples can be more effective than large volumes of lower-quality data. However, the original CoT output from reasoning models often contains substantial redundant information, directly increasing training data size and computational costs for distillation~\citep{qu2025efficientsurvey, luo2025deconstructing}.

% \textbf{Reducing Reasoning Costs via Budget Control.}

% To address the verbosity problem in LLM reasoning, researchers have developed various ``budget control'' techniques~\citep{s1}. These strategies are diverse, with most focusing on inference-time optimizations rather than training efficiency. One approach involves prompt engineering and reasoning time interventions to guide LLMs in generating shorter or more structured reasoning paths without retraining the model~\citep{aa}. Examples include Chain of Draft~\citep{xu2025cod}
% for concise intermediate steps, cognitively inspired ``sketches'' in Sketch-of-Thought~\citep{sot}, explicit tag budgets in hints~\citep{han2024token}, and iterative summarization~\citep{InftyThink} for managing long contexts. Another category enforces budget control through training, with methods like LightThinker~\citep{jin2025lightthinker} that compresses ideas into gist ``tokens'', reinforcement learning approaches to control output length~\citep{pranjal2025L1, shortbetter}, and adaptive reasoning frameworks like AdaR1~\citep{luo2025adar1} that learn to switch between long and short CoT styles. In contrast, research on optimizing the efficiency of training data itself remains significantly underexplored~\citep{luo2025deconstructing}, creating a critical gap that limits the widespread adoption of advanced reasoning capabilities in smaller models.

\textbf{Reasoning Pruning via Budget Control.}
To mitigate this, researchers have developed various ``budget control'' techniques to control the length of CoT output. However, most existing strategies mainly focus on inference-time optimizations~\citep{s1, teng2025atom}, such as LightThinker~\citep{jin2025lightthinker},Sketch-of-Thought~\citep{sot} and AdaR1~\citep{luo2025adar1}. In contrast, training efficiency remains relatively underexplored. Existing methods for this purpose fall into two main categories: (1) Learning-based approaches~\citep{yu2025rethinking, zhao20251} that train auxiliary models to produce shorter reasoning traces, which can adaptively switch between Long and Short CoTs but often incur substantial additional training cost. (2) Prompt engineering-based pipelines that heuristically rewrite CoTs without training extra models. Examples include DLCoT~\citep{luo2025deconstructing}, which introduces a structured framework (DLCoT) to segment, simplify, and optimize Long CoT traces via prompt-based rewriting, and Long-Short Chain-of-Thought Mixture Supervised Fine-Tuning~\citep{yu2025long}, which combines Long and structure-preserved Short CoTs to train models that balance reasoning depth and efficiency. However, to the best of our knowledge, no prior work has explored using difficulty-aware prompting to dynamically adapt CoT length based on problem complexity, leaving a critical gap in scalable and cost-effective CoT distillation.

\section{LiteCoT Dataset: Concise Reasoning with Big Gains}
\label{sec:Methods}

% \begin{figure}[t!]
%     \centering
%     \includegraphics[width=\textwidth]{figures/intro-fig2.pdf}
%     \vspace{-1em}
%     \caption{(a) The training loss comparison chart shows the training loss changes of the four model configurations, Base Long, Base Short, Instruct Long, and Instruct Short, over 90 training steps. (b) The evaluation loss comparison chart shows the corresponding evaluation loss performance of the same four model configurations. (c) The difference in output tokens between ShortCoT and LongCoT distilled models on different benchmarks. \yuyu{ShortCoT?? or LiteCoT??}}
%     \label{fig:motivation_loss_curve}
% \end{figure}

\subsection{Difficulty-Aware Prompting for Curating LiteCoT}
\label{DAP Pipeline}

% We formalize the standard CoT distillation process and then describe our difficulty-aware extension.

We now describe our difficulty-aware prompting (DAP) pipeline, designed to construct a concise and difficulty-adaptive distilled reasoning dataset. The pipeline consists of two sequential steps: (1) generating initial long reasoning solutions, and (2) refining these chains using difficulty-aware prompts to produce shorter, complexity-aligned reasoning traces.

\textbf{Step 1 (Long CoT Generation).}
In conventional distillation, a teacher LLM $M_{teacher}$ is used to generate a reasoning chain $CoT_{L,i}$ for each query $Q_i$ in the training set $D$. Formally:
$CoT_{L,i} = M_{teacher}(Q_i)$ for $i=1,2,\dots,N$. The collected set of long chains is $C_L = {(Q_i, CoT_{L,i})}_{i=1}^N$. These Long CoTs often contain redundant content and do not adjust in length for query difficulty.

\textbf{Step 2 (Difficulty-aware CoT Refinement).} In this step, the teacher model takes each $(Q_i, CoT_{L,i})$ pair as input, along with a specially designed prompt $P_{DA}$ that instructs it to produce a shorter solution appropriate for the difficulty of $Q_i$. This prompt $P_{DA}$ provides different guidelines depending on whether a question is easy, medium, or hard. For example, it might include: ``For easy questions: provide a brief solution$\ldots$ For hard questions: include all necessary detailed steps $\ldots$''. Guided by $P_{DA}$, the teacher first assesses the difficulty of the problem and then refines the solution accordingly:

\begin{enumerate}[leftmargin=2em, topsep=2pt, itemsep=2pt]
    \item \textbf{Difficulty Assessment}:
    Guided by $P_{DA}$, the model first assesses the difficulty of the input query $Q_i$, classifying it into a predefined set of difficulty levels $\mathcal{K} = \{k_{\text{easy}}, k_{\text{medium}}, k_{\text{hard}}\}$. This assessment process can be represented as:
    \begin{equation}
    d_i = M_{teacher}(Q_i, CoT_{L,i}, P_{DA}) \label{eq:difficulty_assessment_en}
    \end{equation}
    where $d_i \in \mathcal{K}$ is the assessed difficulty level. The prompt $P_{DA}$ contains explicit instructions for different assessments.

    \item \textbf{Difficulty-Adaptive CoT Refinement}:
    Based on the assessed difficulty level $d_i$, the model follows the sub-prompt or structured output template $P_{d_i} \subset P_{DA}$ within $P_{DA}$ designed for that specific difficulty $d_i$. It then prunes, reorganizes, or guides the generation of a new, difficulty-matched short CoT, $CoT_{S,d_i}$. This process can be represented as:
    \begin{equation}
    CoT_{S,d_i} = M_{teacher}((Q_i, CoT_{L,i}), P_{DA}, d_i) \label{eq:adaptive_cot_gen_v1_en}
    \end{equation}
    Or more specifically, if $P_{d_i}$ is an explicit output template:
    \begin{equation}
    CoT_{S,d_i} = M_{teacher}(\text{input}=(Q_i, CoT_{L,i}), \text{template}=P_{d_i}(d_i)) \label{eq:adaptive_cot_gen_v2_en}
    \end{equation}
    where $P_{d_i}(d_i)$ denotes the specific instruction part within $P_{DA}$ for difficulty $d_i$ (\eg ``For easy questions: $\ldots$'', ``For medium questions: $\ldots$'', ``For hard questions: $\ldots$''). 
\end{enumerate}

% \begin{figure}[t!]
%     \centering
%     \includegraphics[width=\textwidth]{figures/intro-fig2.pdf}
%     \vspace{-1em}
%     \caption{(a) The training loss comparison chart shows the training loss changes of the four model configurations, Base Long, Base Short, Instruct Long, and Instruct Short, over 90 training steps. (b) The evaluation loss comparison chart shows the corresponding evaluation loss performance of the same four model configurations. (c) The difference in output tokens between ShortCoT and LongCoT distilled models on different benchmarks. \yuyu{ShortCoT?? or LiteCoT??}}
%     \label{fig:motivation_loss_curve}
% \end{figure}

Applying this process to all training samples yields a new distilled dataset
$C_S = {(Q_i, CoT_{S,i})}{i=1}^N$ comprised of shortened, difficulty-graded reasoning traces. 
% Importantly, as illustrated by the bar chart on the right in Figure~\ref{fig:motivation_loss_curve}, the lengths of the refined solutions now correlate with difficulty: solutions for easy queries are the shortest, and those for hard queries are the longest. 
In addition, the refined CoTs are much shorter than the originals (for easy and medium queries, $Length(CoT{S,i}) \ll Length(CoT_{L,i})$). This adaptive reduction strategy is expected to significantly improve the efficiency of model training while preserving essential reasoning for challenging tasks. Ultimately, a student model trained on $C_S$ should learn to solve problems with an appropriate amount of reasoning. 

\subsection{LiteCoT Overview}
\begin{wraptable}{r}{0.55\textwidth}
    \vspace{-\intextsep} % 稍微调整一下垂直间距，可选
    \centering % 使minipage在其分配的空间内居中
    \begin{minipage}{\linewidth} % minipage的宽度现在是相对于wraptable的宽度
        \captionof{table}{Performance comparison of different budget control methods. ``$\Delta$ Tokens'' denotes average tokens per CoT sample.} % 使用 \captionof
        \label{tab:datasets} % 建议使用新的label以区分
        % \small
        \begin{tabular}{lcc}
            \toprule
            Datasets & \#-Samples  & $\Delta$ Tokens  \\
            \midrule
            Light-R1~\cite{lightr1}       & 75.9K & 6382 \\
            OpenThoughts~\cite{open_thoughts}    & 113.9K  & 6801 \\
            OpenThoughts2~\cite{open_thoughts}   & 1M      & 5021 \\
            S1~\cite{s1}             & 1K      & 5122 \\
            LIMO~\cite{ye2025limo}           & 817     & 6984 \\
            SKY-T1~\cite{sky_t1_2025}         & 16K     & 4095 \\
            ReasonFlux~\cite{yang2025reasonflux}     & 1K  & 10500 \\\hline
            \textbf{LiteCoT} (ours)  & 100K    & \textbf{720} \\
            \bottomrule
        \end{tabular}
    \end{minipage}
    \vspace{-\intextsep} % 稍微调整一下垂直间距，可选
\end{wraptable}

\begin{figure}[t!]
    \centering
    \includegraphics[width=\textwidth]{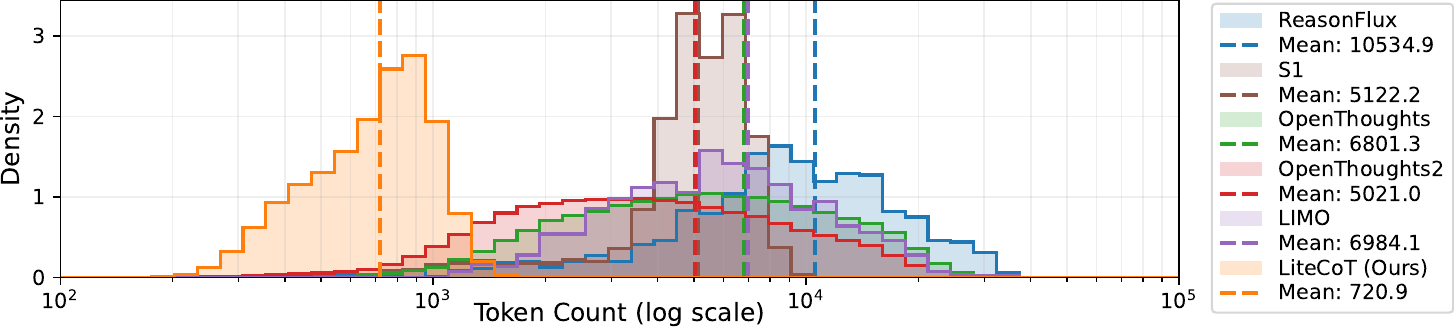}
    \vspace{-1.5em}
    \caption{Token count distribution of LiteCoT compared with previous CoT distillation datasets. }
    \label{fig:token_cnt_distr}
    \vspace{-1.em}
\end{figure}

% Leveraging the DAP methodology detailed in Section~\ref{DAP Pipeline}, we have constructed a new dataset, which is highly efficient for training LLMs. Table~\ref{tab:datasets} presents a comparison of our DAP Pipeline-generated dataset against several other existing CoT datasets, focusing on the number of samples and the average number of tokens per sample ($\Delta$ Tokens).

\textbf{LiteCoT Overview.}
Using the DAP pipeline, we created a distilled dataset, named LiteCoT, of 100K question–solution pairs with difficulty-adaptive CoTs. 
Table~\ref{tab:datasets} highlights the token savings of our approach compared to prior distillation data. Other budget-control methods still produce CoT solutions averaging several thousand tokens each (\eg 5K–10K tokens for ReasonFlux~\citep{yang2025reasonflux} and OpenThoughts~\citep{open_thoughts}), often necessitating tens or hundreds of thousands of training samples. In stark contrast, LiteCoT's CoT samples average only about 720 tokens, representing an order-of-magnitude improvement in token efficiency.

Figure~\ref{fig:token_cnt_distr} clearly shows this advantage: the token-count distribution for LiteCoT is markedly shifted toward shorter reasoning chains compared to other datasets. 
% This concise and structured distribution not only significantly reduces computational and training overhead but also allows student models to train more efficiently with fewer tokens, potentially enabling a broader exploration of diverse reasoning examples within the same token budget. 
% As demonstrated in subsequent experiments, this substantial token reduction is achieved without sacrificing reasoning performance and often even enhances the performance of distilled models.
Please refer to Appendix~\ref{appen:dataset examples} for details. 

\textbf{Training the \model with \data.}
Using LiteCoT, we distilled a new family of reasoning models, named \texttt{Liter} (1.5B, 7B, and 32B), based on the Qwen2.5. We followed the training strategy from Open-R1~\citep{openr1} and fine-tuned the Liter models for 3 epochs on the LiteCoT dataset. Specifically, we employed a learning rate of 5e-5 with a cosine learning rate scheduler, including a 10\% warm-up phase. In addition, we increased the RoPE frequency to 100K, effectively extending the context length from 4K to 8K tokens.
Please refer to Appendix~\ref{appendix:experiment details} for more training or evaluation details.

\section{Experiments}
\label{sec:Experiments}
We conduct extensive experiments to evaluate the effectiveness of LiteCoT data.
Our evaluations address three key questions:
(\textbf{Exp-1}) Does training on short difficulty-aware CoTs, LiteCoT, actually improve model performance compared to long CoTs?
(\textbf{Exp-2}) Can our DAP method rival or surpass models distilled with an order of magnitude more data from a powerful teacher? 
(\textbf{Exp-3}) How does our DAP approach compare to other recent CoT distillation and compression techniques?

% \yuyu{where is the settings of \model model???}

% In the experiments section, we want to conduct several experiments to answer the following questions:
% \begin{itemize}[leftmargin=2em, topsep=2pt, itemsep=2pt]
%     \item Q1: Can the short CoT data produced by the DAP pipeline surpass the long CoT data? (Exp-1)
%     \item Q2: Can we surpass the official distillation model of DeepSeek R1 with more training data? (Exp-2)
%     \item Q3: Does the data obtained based on the DAP pipeline have significant advantages over other current CoT decompose methods? (Exp-3)
%     % \item Q4: What role does short reasoning data play in reinforcement learning?
% \end{itemize}

\subsection{Experiment Setup}
\stitle{Training Data.} 
Our experimental evaluation consists of three stages. All data comes from the OpenThoughts~\cite{open_thoughts} dataset and is processed using our DAP distillation process. For \textbf{Exp-1}: Use 25$K$ OpenThoughts samples. Their original Long CoT form is used as the baseline training data; at the same time, these samples are processed by the DAP process to generate 25$K$ corresponding Short CoT data for comparison training. For \textbf{Exp-2}: Expand the amount of OpenThoughts data to 100$K$. These samples are processed by the DAP process to generate 100$K$ short CoT data(\ie LiteCoT), which constitute the training set of this experiment. For \textbf{Exp-3}: Directly use the 25$K$ short CoT data generated in \textbf{Exp-1} as the training data for this stage.

\stitle{Base Models.} In \textbf{Exp-1} and \textbf{Exp-3}, due to the small amount of training data, the models we selected are all instruction versions in order to obtain better instruction following capabilities on small datasets.
For \textbf{Exp-1}: Qwen2.5-0.5B-Instruct, Qwen2.5-Math-1.5B-Instruct, Qwen2.5-Math-7B-Instruct, Qwen2.5-14B-Instruct of the Qwen2.5 series, and Llama3.1-8B-instruct of the Llama3.1 series are included as base models. For \textbf{Exp-2}: Aiming to make a fair comparison with the results in the DeepSeek R1 report, we selected the same base version models used in the report: Qwen2.5-math-1.5B, Qwen2.5-math-7B, Qwen2.5-14B, and Qwen2.5-32B. For \textbf{Exp-3}: Similar to {Exp-1}, the model Qwen2.5-math-7B-Instruct is selected as the baseline model.

\stitle{Benchmark Datasets.} We evaluate model accuracy on a diverse suite of 11 benchmarks, covering math word problems (MATH500~\cite{hendrycks2021measuring}, GSM8K~\cite{cobbe2021training}), competition-level math (AIME24~\cite{aime2024}/AIME25~\cite{aime2025},OlympiadBench~\cite{choquette2023correlated}, MinervaMath~\cite{minerva_math}), general QA (GPQA)~\cite{rein2024gpqa}, and academic exams (GaoKao2023~\cite{zhang2023evaluating}, MMLU-STEM~\cite{MMLU_STEM}, etc.).

% Our three experiments include the following benchmarks: Math500, GPQA, AIME, AMC, GSM8K, Minerva Math, GaoKao Math, College Math, SAT Math, MMLU-STEM. We chose these benchmarks for evaluation because they not only comprehensively examine the performance of models on different levels of mathematical reasoning tasks and a variety of basic subject knowledge, but are also the most common and core evaluation tools for RLMs. 

\stitle{Baselines.} For \textbf{Exp-1}, we mainly use the model directly distilled from the original Long CoT as the baseline to evaluate the relative improvement of our method. For \textbf{Exp-2}, we include a series of current mainstream reasoning models in the benchmark comparison range, including representative models such as S1~\cite{s1} and OpenThoughts~\cite{open_thoughts}. 
For \textbf{Exp-3}, our benchmark comparison focuses on different Long CoT optimization or ``decompose'' methods. These methods include strategies that optimize at the prompt engineering level, such as Chain-of-Draft~\cite{xu2025cod} and Budget Aware Reasoning~\cite{han2024token}; in addition, we also include methods that require training additional models to simplify the output content, such as LLMLingua2~\cite{pan2024llmlingua}.

\begin{table}[t!]
    \centering
    \small
    % \vspace{-1em}
    \caption{Performance comparison of Long CoT distilled models and short CoT distilled models. \colorbox{green!10}{Green} means short CoT distilled models outperform the corresponding Long CoT distilled models. \colorbox{pink!20}{Pink} means the opposite (``QW'' and ``LM'' refer to Qwen2.5-Instruct and Llama3.1-Instruct).}
    % \vspace{-1em}
    \label{tab:longvsshort}\small
    \begin{tabular}{l|cccccccccccccc} 
        \toprule
         \multirow{2}{*}{\textbf{Benchmark}}
         % & \multicolumn{2}{c}{PPL$\downarrow$ }
         & \multicolumn{2}{c}{QW-0.5B} 
         & \multicolumn{2}{c}{QW-1.5B} 
         & \multicolumn{2}{c}{QW-3B} 
         & \multicolumn{2}{c}{QW-7B} 
         & \multicolumn{2}{c}{QW-14B}
         & \multicolumn{2}{c}{LM-8B} \\
         & \scriptsize short& \scriptsize long
         % & \scriptsize short& \scriptsize long 
         & \scriptsize short& \scriptsize long 
         & \scriptsize short& \scriptsize long 
         & \scriptsize short& \scriptsize long 
         & \scriptsize short& \scriptsize long  
         & \scriptsize short& \scriptsize long \\
        \midrule
        % Perplexity$\downarrow$& \textbf{1.48} & 2.00 & \textbf{1.38} & 1.71 & \textbf{1.34} & 1.60 & \textbf{1.33} & 1.60 & & &\textbf{1.37} &1.59\\ 
        MATH500& \cellcolor{green!10}18.6 & 11.6 & \cellcolor{green!10}62.8 & 56.4 & \cellcolor{green!10}47.8 & 46.0 & \cellcolor{green!10}66.2 & 64.8 & \cellcolor{green!10}65.0 & 64.4& \cellcolor{green!10}34.2 & 32.2\\
        AIME24& 0.0 & 0.0 & \cellcolor{green!10}3.3 & 0.0 & \cellcolor{green!10}3.3 & 0.0 & \cellcolor{green!10}23.3 & 3.3 &\cellcolor{green!10}13.3 & 10.0 &\cellcolor{green!10}3.3 &0.0\\
        GPQA & \cellcolor{green!10}22.7 & 21.2 & \cellcolor{green!10}28.8 & 27.3 & \cellcolor{green!10}23.7 & 22.7 & \cellcolor{pink!20}29.8 & 32.3 & \cellcolor{green!10}40.4 & 38.4 & \cellcolor{green!10}32.3 & 25.3\\
        GSM8K& \cellcolor{green!10}33.6 & 20.5 & \cellcolor{green!10}85.1 & 84.4 & \cellcolor{green!10}77.3 & 61.4 & \cellcolor{green!10}95.5 & 95.0 & \cellcolor{green!10}90.8 & 85.1 & \cellcolor{green!10}73.7 & 26.8\\ 
        Minerva Math& \cellcolor{pink!20}3.7 & 4.4 & \cellcolor{green!10}25.4 & 19.5 & \cellcolor{green!10}14.3 & 11.8 & \cellcolor{green!10}33.9& 31.2 & \cellcolor{green!10}29.0 & 21.3 &\cellcolor{green!10}10.3 & 7.7\\
        GaoKao2023 & \cellcolor{green!10}23.4 & 11.9 & \cellcolor{green!10}65.2 & 62.9 & \cellcolor{green!10}43.3 & 39.5 & \cellcolor{green!10}66.5 & 66.2 & \cellcolor{green!10}56.9 & 49.9 & 
        \cellcolor{green!10}32.2 & 25.2\\
        OlympiadBench & \cellcolor{green!10}5.9 & 1.5 & \cellcolor{green!10}37.2 & 36.3 & \cellcolor{green!10}15.4 & 10.8& \cellcolor{green!10}42.8 & 37.5 & \cellcolor{green!10}28.4 & 19.1 & \cellcolor{green!10}13.0 & 11.6 \\
        College Math & \cellcolor{green!10}19.6 & 9.8 & \cellcolor{pink!20}46.9 & 47.2 & \cellcolor{green!10}35.6 & 31.2 & \cellcolor{green!10}48.5 & 47.5 & \cellcolor{green!10}41.9 & 32.1 & \cellcolor{green!10}24.7& 1.61\\
        Sat Math& \cellcolor{green!10}37.5 & 12.5 & \cellcolor{green!10}43.8&3.1 &\cellcolor{green!10}68.8 & 34.4 & \cellcolor{green!10}96.9 & 90.6 & \cellcolor{green!10}96.9 & 62.5 &\cellcolor{green!10}43.8 & 25.0\\
        MMLU STEM& \cellcolor{green!10}29.7 & 9.2 & \cellcolor{green!10}33.9 & 10.4 & \cellcolor{green!10}58.1 & 21.4 & \cellcolor{green!10}64.4 & 60.5 & \cellcolor{green!10}71.3 & 56.9 & \cellcolor{green!10}33.9 & 29.0\\
        ACM23& \cellcolor{green!10}2.5 &0.0 &\cellcolor{green!10}47.5 &45.0 &\cellcolor{green!10}30.0 &17.5 &\cellcolor{green!10}62.5&57.5 & \cellcolor{green!10}42.5 & 27.5 &\cellcolor{pink!20}5.0 &25.0 \\ 
        \midrule
        Overall & \cellcolor{green!10}17.9 & 9.3 & \cellcolor{green!10}43.6 & 35.7 & \cellcolor{green!10}38.0 & 27.0 & \cellcolor{green!10}57.3 & 53.3 & \cellcolor{green!10}52.4 & 42.5 & \cellcolor{green!10}27.9 & 19.0 \\ % \midrule 
        \bottomrule
    \end{tabular}
\end{table}

\subsection{Experimental Results and Analysis}
% \yuyu{Can we summarize some interesting findings or conclusions within each Exp? If can't, we can remove the below box.}

% \begin{tikzpicture}
% \node [mybox] (box){%
%     \begin{minipage}{\columnwidth}
%     {
%     Findings here ......
%     }
%     \end{minipage}
% };
% \node[fancytitle, rounded corners] at (box.north) {\small{\bf Takeaway 1}};
% \end{tikzpicture}

\subsubsection{Exp-1: LiteCoT vs. Long CoT Baselines} We evaluate the fine-tuned models trained on our \data and Long CoT data, comparing their overall performance and analyzing the accuracy-speed trade-off.

\textbf{Overall Performance.} As shown in Table~\ref{tab:longvsshort}, we comprehensively compare the average accuracy of Qwen2.5 series models and Llama3.1-8B trained with Short CoT and Long CoT on 11 different benchmarks. The average performances show that all models trained and distilled with short CoT data significantly outperform the corresponding models distilled with Long CoT data. For example, the overall average performance of Qwen2.5-1.5B-Instruct (short) reached 43.6\%, while its Long CoT version was 35.7\%; Qwen2.5-7B-Instruct (short) reached 57.3\%, higher than the 53.3\% of the Long CoT version. This trend is consistently reflected in models of different parameter sizes.

\begin{figure}[t!]
    \centering
    \vspace{-1em}
    \includegraphics[width=\textwidth]{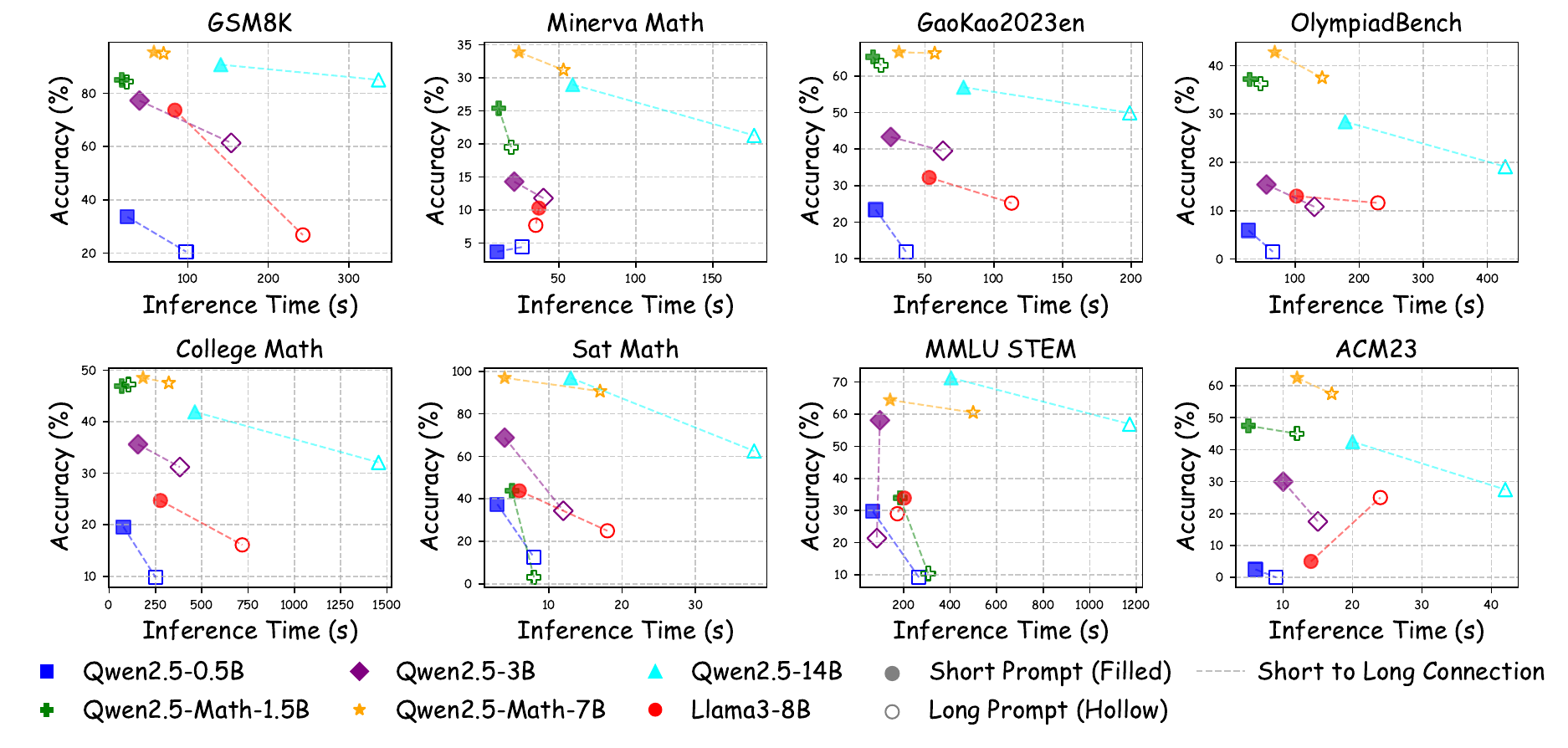}
    \vspace{-2em}
    \caption{Inference time vs. accuracy across eight benchmarks. Models trained with short CoT (filled symbols) demonstrate significantly faster inference times compared to Long CoT (hollow symbols) while maintaining or exceeding accuracy levels.}
    \label{fig:acc_vs_inference_time_on_long_and_short}
    \vspace{-1em}
\end{figure}

\textbf{The relationship between inference speed and accuracy.} Figure~\ref{fig:acc_vs_inference_time_on_long_and_short} clearly shows that for all models (Llama3.1-8B and Qwen2.5 series models), the use of Short CoT (filled points) generally achieves inference time that is much lower than Long CoT (hollow points). More importantly, this significant speed improvement does not come at the expense of accuracy; on the contrary, in most benchmarks, Short CoT is not only faster, but its accuracy can also reach or even exceed the level of Long CoT. For example, on multiple benchmarks such as GSM8K and MMLU STEM, Short CoT data points all show a better ``speed-accuracy'' balance. This finding spans models of different parameter sizes and a variety of task types, and strongly proves that the DAP Short CoT method can significantly improve inference efficiency while maintaining or even optimizing the model's inference accuracy.
\stitle

\begin{table}[htbp]
    \centering
    \small
    \caption{Performance Comparison of Language Models on Reasoning Benchmarks. Since there are too many models to be tested, we only select some of them for retesting, and we directly copy the results of all other models from their papers or blogs. $\dagger$ represents the copied results, $\ddagger$ represents the results retested by ourself. DS-R1-Distill represents the official Distilled Models in the DeepSeek-R1 report. OB represents the Olympiad Bench. \textbf{Bold} means the first performance and \underline{underline} means the second performance.}
    % \vspace{-1em}
    \label{tab:modelcomparison_of_the_same_parameter}\small
    \begin{tabular}{l|cccccc}
	\toprule
	\textbf{Model} & \textbf{AIME24} & \textbf{MATH500} & \textbf{GPQA} & \textbf{AMC23} & \textbf{OB} & \textbf{OVERALL} \\
	\midrule
	\rowcolor{gray!15}\multicolumn{7}{c}{\textbf{Frontier Models}} \\
	\midrule
	GPT-4o$^\dagger$ &9.3 & 76.6 & 56.1 &47.5 & \underline{43.3} &46.56\\
	Claude3.5-Sonnet$^\dagger$ &16.0 & 78.1 & 59.4& 44.6 & - & - \\
	GPT-o1-preview$^\dagger$ &\underline{44.6} & \underline{85.5} & \textbf{73.3} & \underline{90.0} & - & - \\
	GPT-o1-mini$^\dagger$ &\textbf{56.7} &\textbf{90.0} & \underline{60.0}& \textbf{95.0} & \textbf{65.3}& \textbf{73.4} \\
	\midrule
	\rowcolor{gray!15}\multicolumn{7}{c}{\textbf{Open-Sourced Reasoning LLMs}} \\
	\midrule
	DeepSeek-V3$^\dagger$ & 39.2 & \underline{90.2} & \underline{59.1} & \textbf{85.0} & 55.4 & 65.78 \\
	Mathstral-7B-V0.1$^\dagger$ & 0.0& 57.8 & -& 37.5 & 21.5 & - \\
	DS-Coder-V2-Instruct$^\dagger$ &13.3 & 75.3 & - & 57.5&37.6 & - \\
	Qwen2.5-Math-72B-Instruct$^\dagger$ & 30.0 & 85.6 & - & 70.0 & \underline{49.0} & - \\
	NuminaMath-72B-CoT$^\dagger$ & 3.3 & 64.0 & - & 70.0 & 32.6 & - \\
	LLaMA3.1-70B-Instruct$^\dagger$ & 23.3 & 65.4 & 46.7 & 50.0 & 27.7 & \textbf{42.62} \\
	LLaMA3.1-405B-Instruct$^\dagger$ & - & 73.8& 51.1 & - & 34.8 & - \\
	DS-Distilled-LLaMA3.1-70B$^\dagger$ & \textbf{70.0} & \textbf{94.5} & \textbf{65.2} & - & - & - \\
	rStar-Math$^\dagger$ & \underline{43.3} & 88.2 & - & \underline{80.0} & \textbf{63.1} & - \\
	\midrule
	\rowcolor{gray!15}\multicolumn{7}{c}{\textbf{1.5B}} \\
	\midrule
	Qwen2.5-Math-1.5B$^\ddagger$ & 0.0 & 51.2 & 0.0 & 22.5 & 16.7 & 18.08 \\
	Qwen2.5-Math-1.5B-instruct$^\dagger$ & 10.0 & 60.0 & - & 60.0 & 38.1 & - \\
	ReasonFlux-1.5B$^\dagger$ & 20.0 & 70.4 & - & \textbf{72.5} & \textbf{49.0} & - \\
	DS-R1-Distilled-1.5B$^\ddagger$ & \underline{28.9} & \underline{83.9} & \underline{33.8} & 55.0 & 31.4 & 46.6 \\
    \hline
    \rowcolor[gray]{.8}
	\textbf{\model-1.5B}$^\ddagger$ & \textbf{36.7} & \textbf{87.2} & \textbf{35.4} & \underline{60.0} & \underline{42.7} & \textbf{52.4} \\
	\midrule
    \rowcolor{gray!15}\multicolumn{7}{c}{\textbf{7B}} \\
     \midrule
	Qwen2.5-Math-7B$^\ddagger$ &10.0 & 74.2 & 33.3 & 45.0 & 21.8 & 36.86 \\
	Qwen2.5-math-7B-instruct$^\dagger$ & 13.3 & 82.6 & - & 62.6 & 39.0 & - \\
	Openthinker-7B$^\dagger$ & 31.3 & 83.2 & 42.9 & 74.5 & - & - \\
	Openthinker2-7B$^\dagger$ & 50.0 & 88.4 & \underline{49.3} & \textbf{89.5} & - & - \\
	Light-R1-7B$^\dagger$ & 53.3 & \underline{93.2} & 46.5 & 85.0 & \underline{52.3} & 66.06 \\
	OpenR1-Qwen-7B$^\dagger$ & 48.7 & 87.8 & 21.2 & \underline{88.5} & - & - \\
	DS-R1-Distilled-7B$^\ddagger$ &\textbf{55.5} & 92.8 & 49.1 & 77.5 & 49.8 & 64.94 \\
	Bespoke-Stratos-7B$^\dagger$ & 20.0 & 82.0 & 37.8 & - & - & - \\
	OlympicCoder-7B$^\dagger$ & 20.7 & 74.8 & 25.3& 63.0 & - & - \\
	ReasonFlux-7B$^\dagger$ & 36.7 & 88.6& - & 80.0 & \textbf{54.8} & - \\
    \hline
    \rowcolor[gray]{.8}
	\textbf{\model-7B}$^\ddagger$ & \underline{53.3} &\textbf{93.8} & \textbf{53.3} & 82.5 & 50.2 & \textbf{66.62} \\
	\midrule
	\rowcolor{gray!15}\multicolumn{7}{c}{\textbf{32B}} \\
	\midrule
	Qwen2.5-32B-instruct$^\dagger$ &13.3 & 79.4 & 33.8 & 37.5 & - & - \\
	Openthinker-32B$^\dagger$ & 68.0 & 90.6 & 63.5 & - & - & - \\
	Openthinker2-32B$^\ddagger$ & 63.3 & 96.0 & \textbf{65.1} & 80.0 & \underline{56.3} & 72.14 \\
	DS-R1-Distilled-32B$^\ddagger$ & \underline{72.6} & 95.2 & 62.1 & \textbf{90.0} & 53.0 & 74.58 \\
	Light-R1-32B$^\dagger$ & 70.0 & 94.2 & 61.1 & 77.5 & - & - \\
	LIMO-32B$^\ddagger$ & 40.0 & 95.0 & 51.0 & 52.5 & 43.9 & 56.48 \\
	S1.1-32B$^\ddagger$ & 43.3 & 89.2 & 52.5 & 75.0 & \textbf{56.6} & 63.32 \\
	ReasonFlux-32B$^\ddagger$ & 73.3 & \underline{96.4} & 63.1 & 70.0 & 34.7 & 67.5 \\
	Bespoke-Stratos-32B$^\dagger$ & 63.3 & 93.0 & 58.1 & - & - & - \\
	Sky-T1-32B-preview$^\dagger$ &43.3 & 82.4 & 56.8 & - & - & - \\
	QwQ-32B-Preview$^\ddagger$ & 76.7 & 92.2& 65.1 & 72.5 & 54.1 & 72.12 \\
    \hline
    \rowcolor[gray]{.8}
	\textbf{\model-32B}$^\ddagger$ & \textbf{76.7} & \textbf{96.6} & \underline{63.6} & \underline{87.5} & 53.9 & \textbf{75.66} \\
	\bottomrule
 \end{tabular}
\end{table}

\subsubsection{Exp-2: Comparing our Liter Models with Mainstream Distilled LLMs} 
% We first compared the trained model with the model distilled from DeepSeek-R1 Reported. Then, to verify the effectiveness of our distilled models, we compared the distilled 7B and 32B models with almost all the mainstream inference models on multiple benchmarks.
We first compared our Liter models against models distilled from DeepSeek-R1 (\textbf{Exp-2.1}).
To further evaluate the effectiveness of our \texttt{Liter} models, we compared our Liter 7B and 32B variants against a broad range of mainstream distilled models across multiple reasoning benchmarks (\textbf{Exp-2.2}).

\begin{figure}[!t]
    \centering
    \includegraphics[width=\textwidth]{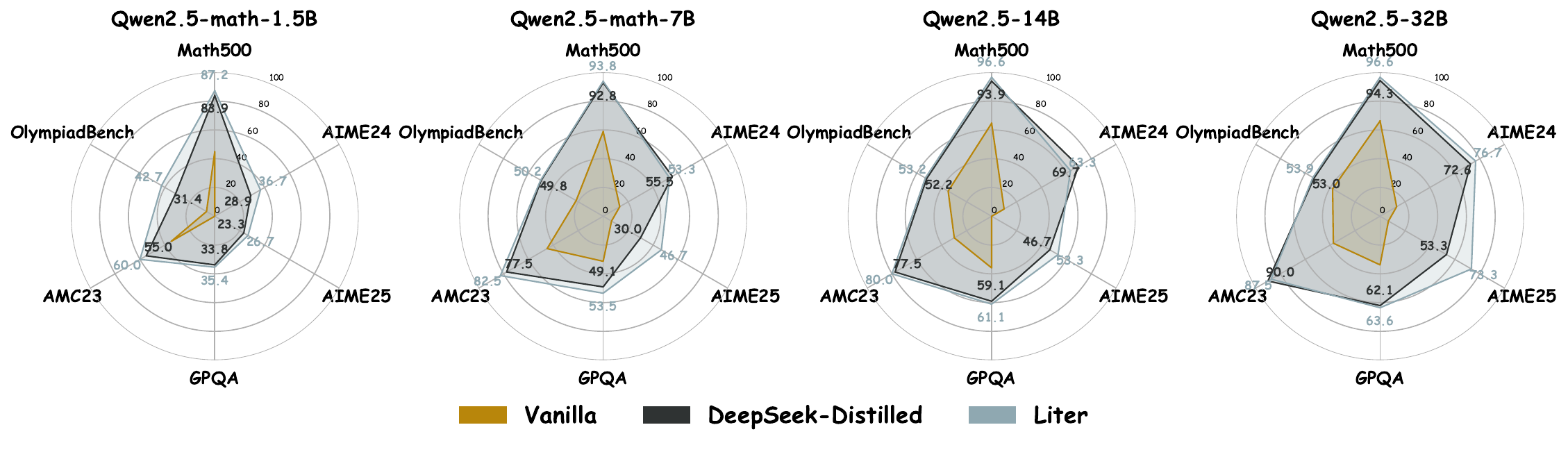}    
    \vspace{-1.5em}
    \caption{Radar chart show the performance comparison of the Vanilla, DeepSeek-R1-Distilled, and Liter versions of the Qwen2.5 model with different parameter sizes (1.5B, 7B, 14B, 32B) on multiple math and reasoning benchmarks (Math500, AIME24, OlympiadBench, AIME25, GPQA, AMC23).}
    \label{fig:radar_curve_ds_r1_vs_ours}
    \vspace{-1em}
\end{figure}

\textbf{Exp-2.1: Comparison with DeepSeek-R1-Distilled Models.} Figure~\ref{fig:radar_curve_ds_r1_vs_ours} shows the performance of four models of different parameter sizes (1.5B, 7B, 14B and 32B) in the Qwen2.5 series on six key benchmarks after three training methods: original weights (Vanilla), DeepSeek-R1 distilled models (DeepSeek-R1-Distilled) and our Liter. It shows that the performance of the model representing Liter (Gray blue area) is consistent and significantly better than the other two methods at all parameter scales. Specifically, taking the Qwen2.5-7B model as an example, our method achieved an accuracy of 60.6\% on Math500, while DeepSeek-R1 distilled and the original model achieved 55.8\% and 42.6\% respectively; on the more challenging OlympiadBench, our model scored 30.2\%, also ahead of DeepSeek R1 distilled's 21.1\% and the original model's 10.0\%. This performance advantage is particularly evident in the Qwen2.5-32B model with a larger number of parameters. For example, on the AIME24 benchmark, the accuracy of the DAP distilled model is as high as 76.7\%, far exceeding the 48.0\% of DeepSeek R1 distillation and the 13.3\% of the original model. These specific numerical comparisons strongly prove that the model trained on the LiteCoT can achieve comprehensive performance superiority over other training methods on multiple core benchmarks

\textbf{Exp-2.2: Comparison with Mainstream RLMs of Similar Parameter Sizes.} 
We conducted an extensive horizontal comparison involving our Liter models (1.5B, 7B, and 32B) against various publicly available mainstream reasoning models of comparable parameter scales. Evaluations were carried out on five representative benchmarks: AIME24, MATH500, GPQA, AMC23, and OlympiadBench, with results presented in Table~\ref{tab:modelcomparison_of_the_same_parameter}.

Overall, whether at the parameter scale of 1.5B, 7B, or 32B, the models trained by the DAP distillation method show comparable or better performance than the current mainstream or even top reasoning models. At the 1.5B level, Ours-1.5B achieved an accuracy of 87.2\% on MATH500, significantly better than most of the same-level comparison models, and showed strong competitiveness on OlympiadBench (42.7\%). Entering the 7B level, Ours-7B performs particularly well, achieving an accuracy of 93.8\% on MATH500, and leading the other 7B parameter models listed in the table with 53.3\% on AIME24 and GPQA respectively, while also being at the top level on OlympiadBench (50.2\%). When the model scale is expanded to 32B, the performance advantage of Ours-32B is further consolidated, for example, it ranks among the top on multiple difficult benchmarks such as MATH500 (96.6\%), AIME24 (76.7\%), and GPQA (63.6\%).

% % 换成雷达图
% \begin{table}[htbp]
%     \centering
%     % \vspace{-1em}
%     \caption{Performance of the distilled models trained on full short cot data. }
%     % \vspace{-1em}
%     \label{tab:longvsshort}\small
%     \begin{tabular}{l|cccccccccccccc} 
%         \toprule
%          \multirow{2}{*}{\textbf{Benchmark\scriptsize{(Metric)}}}
%          % & \multicolumn{2}{c}{PPL$\downarrow$ }
%          & \multicolumn{2}{c}{QW-0.5B} 
%          & \multicolumn{2}{c}{QW-1.5B} 
%          & \multicolumn{2}{c}{QW-3B} 
%          & \multicolumn{2}{c}{QW-7B} 
%          & \multicolumn{2}{c}{QW-14B}
%          & \multicolumn{2}{c}{LM-8B} \\
%          & \scriptsize short& \scriptsize long
%          % & \scriptsize short& \scriptsize long 
%          & \scriptsize short& \scriptsize long 
%          & \scriptsize short& \scriptsize long 
%          & \scriptsize short& \scriptsize long 
%          & \scriptsize short& \scriptsize long  
%          & \scriptsize short& \scriptsize long \\
%         \midrule
%         Perplexity$\downarrow$& 1.48 & 2.00 & 1.38 & 1.71 & 1.34 & 1.60 & 1.33 & 1.60 & & &1.37 &1.59\\ 
%         MATH500& & & & & & & & & &\\
%         AIME24& & & & &\\
%         GPQA& & & & &\\
%         GSM8K& & & & &\\ 
%         Minerva Math& & & & &\\
%         GaoKao2023 & & & & &\\
%         OlympiadBench & & & & &\\
%         College Math & & & & & \\
%         Sat Math& & & & &\\
%         MMLU STEM& & & & &\\
%         ACM23& & & & & \\ % \midrule  

%         \bottomrule
%     \end{tabular}
% \end{table}

\begin{figure}[t!]
    \centering
    \includegraphics[width=\textwidth]{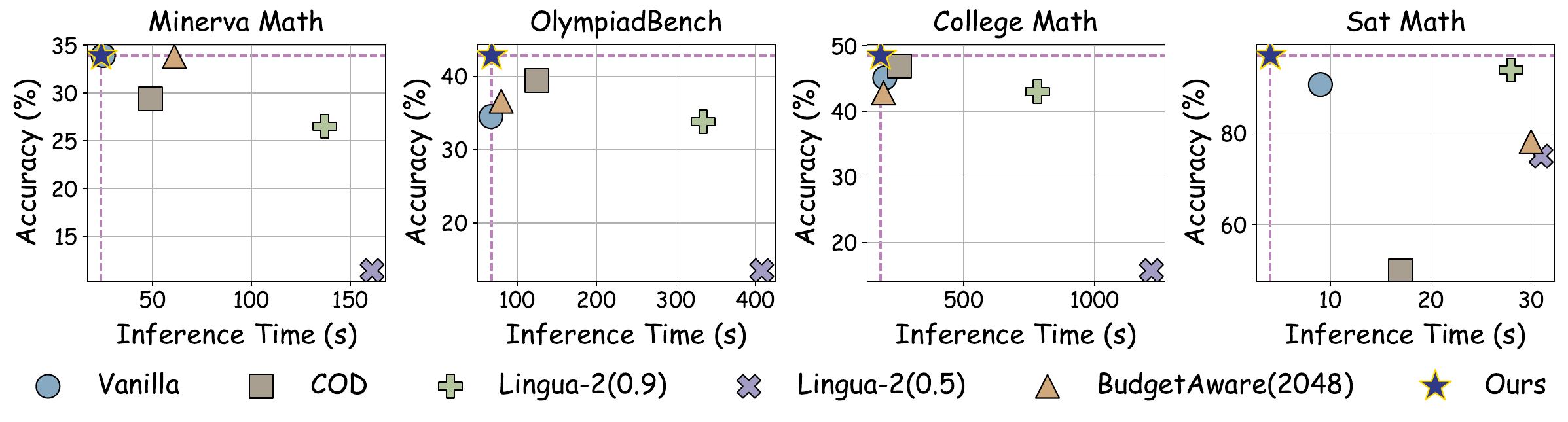}
    \vspace{-1.5em}
    \caption{Scatter plots comparing inference time and accuracy of different methods across multiple benchmarks. Our method (Ours) consistently achieves optimal accuracy while significantly reducing inference time compared to existing methods.}
    \label{fig:different_cot_decomposition_scatter}
\end{figure}

\begin{table}[t!]
    \centering
    % \vspace{-1em}
    \caption{Performance comparison of different CoT decompose methods. COD represents Chain-of-Draft, Lingua-2 means LLMLingua-2.} 
    % \vspace{-1em}
    \label{tab:different_cot_decompose}\small
    \begin{tabular}{l|cccccc}
        \toprule
        \textbf{Benchmark}
        % & \multicolumn{2}{c}{PPL$\downarrow$ }
        & Vanilla
        & COD
        & Lingua-2(0.9)
        & Lingua-2(0.5)
        & BudgetAware(2048)
        & \textbf{Ours} \\

        \midrule

        % MATH500& 62.2 & 76.2 & 76.0 & 35.4 & 75.6 & 66.2 \\
        AIME24& 13.3 & 13.3 & 16.7 & 0.0 & 13.3 & \textbf{23.3} \\
        AIME25& 10.0 & 13.3 & 23.3 & 0.0 & 3.3 & \textbf{16.7} \\
        GPQA & 27.3 & 27.3 & 32.1 & 21.2 & \textbf{34.8} & 29.8 \\
        GSM8K& 94.8 & 94.4 & 94.8 & 72.4 & 94.4 & \textbf{95.5} \\
        Minerva Math& 33.9 & 29.4 & 26.5 & 11.4 & 33.8 & \textbf{33.9} \\
        GaoKao2023en & 64.2 & 62.3 & 64.4 & 34.5 & 65.2 & \textbf{66.5} \\
        OlympiadBench & 34.5 & 39.4 & 33.8 & 13.5 & 36.6 & \textbf{42.8} \\
        College Math & 45.1 & 46.9 & 43.0 & 15.7 & 42.8 & \textbf{48.5} \\
        Sat Math& 90.6 & 50.0 & 93.8 & 75.0 & 78.1 & \textbf{96.9} \\
        MMLU STEM& 68.8 & 42.7 & \textbf{69.5} & 43.7 & 58.9 & 64.4 \\
        ACM23& 60.0 & 45.0 & 45.0 & 27.5 & 52.5 & \textbf{62.5} \\

        \bottomrule
    \end{tabular}
\end{table}

\subsubsection{Exp-3: Comparison with other CoT Optimization methods.} We continue to compare models trained with DAP distilled data against those using other CoT decomposition techniques, focusing on overall performance and the speed-accuracy trade-off to analyze their key differences and advantages.

\textbf{Overall Performance of Different CoT Decomposition Methods.} Table~\ref{tab:different_cot_decompose} clearly shows the performance comparison of different CoT decomposition methods on multiple benchmarks, including the original baseline model (Vanilla), Chain-of-Draft (COD), two LLMLingua-2 with different compression rates (Lingua-2(0.9) and Lingua-2(0.5)), BudgetAware(2048), and our DAP method (Ours). From the accuracy data of each benchmark, our DAP method (Ours) shows the strongest overall performance. For example, on the challenging AIME24 benchmark, Ours achieved an accuracy of 23.3\%, significantly higher than the other methods. On the important SAT Math benchmark, Ours also leads all other comparison methods with an accuracy of 96.9\%, including COD (50.0\%), LLMLingua-2(0.9) (93.8\%), and BudgetAware(2048) (78.1\%). It is worth noting that although some methods such as LLMLingua-2 may show good single-point performance on specific benchmarks (such as Lingua-2(0.9) 32.1\% on GPQA), ours method can maintain or be close to the leading level on most benchmarks.

\textbf{The relationship between inference speed and accuracy.} Figure~\ref{fig:different_cot_decomposition_scatter} fully demonstrates the superiority of the DAP distillation method (labeled as ``Ours'', purple star with gold rim), with additional results shown in Appendix~\ref{appendix:acc_vs_inference_time_on_long_and_short}. In the tests across eight different benchmarks, the ``Ours'' method has the best trade-off in almost all speed-accuracy two-dimensional planes. Specifically, compared with other mainstream CoT decomposition techniques such as Chain-of-Draft, LLMLingua-2, and BudgetAware, our method not only achieves the highest accuracy in most cases (for example, 62.5\% on ACM23 and 96.9\% on Sat Math), but also achieves the shortest or very competitive inference times.
In comparison, other methods either compromise on accuracy (such as LLMLingua-2 (0.5) or COD with high compression ratios) or far exceed ``Ours'' in reasoning time (such as LLMLingua-2 (0.9) with low compression ratios on some tasks). This better ``price/performance ratio'' consistently achieved on multiple benchmarks strongly demonstrates the unique advantages of the DAP distillation method.

\section{Conclusion}

In this paper, we introduced Difficulty-Aware Prompting, a new pipeline for CoT distillation. Leveraging DAP, we created LiteCoT, a difficulty-adaptive CoT dataset consisting of 100K distilled reasoning examples with an average length of only 720 tokens per sample, far shorter than existing alternatives. Utilizing LiteCoT, we distilled and trained a new family of reasoning models named Liter (1.5B, 7B, and 32B), based on the Qwen2.5 architecture. 
Extensive experiments across diverse reasoning benchmarks clearly demonstrate the effectiveness of the LiteCoT dataset and Liter models, consistently outperforming previous approaches that rely on substantially larger datasets with uniformly verbose reasoning chains.

{

\small

\bibliographystyle{IEEEtran} \bibliography{refs/ref.bib}

}

%%%%%%%%%%%%%%%%%%%%%%%%%%%%%%%%%%%%%%%%%%%%%%%%%%%%%%%%%%%%

\appendix
\section*{Appendices}
\section{Difficulty Aware Prompt}
This section provides the complete ``Difficulty Aware Prompt'' used in the study. These prompts are designed to guide the model in assessing the difficulty of a problem and generating reasoning procedures of appropriate length accordingly.

\begin{tcolorbox}[title={\textbf{\small Yaml config for running Qwen2.5-Math-7B on LiteCoT}}, boxrule=2pt, arc=0mm, breakable]\begin{verbatim}
    
As an assistant, you need to first assess the complexity of the problem 
and adopt an appropriate thinking framework before providing the final 
solution. Structure your response into two main sections: Thought and 
Solution.

First evaluate the complexity of the problem, then choose a suitable 
thinking framework, and describe the thought process as detailed as 
possible:

1. For simple problems:

**Analysis:**
[Understand the core elements and goals of the problem]

**Approach:**
[Propose direct solution methods]

**Summary:**
[Concisely summarize the solution approach and key points]


2. For moderately complex problems:

**Analysis:**
[Understand the problem and identify key points and challenges]

**Initial Approach:**
[Propose preliminary solutions]

**Reflection:**
[Evaluate the pros and cons of the initial approach]

**Improvement:**
[Refine the solution based on reflection]

**Summary:**
[Summarize the key points of the final solution]

3. For highly complex problems:

**Analysis:**
[Analyze various aspects of the problem and break down its structure]

**Problem Decomposition:**
[Break complex problem into manageable sub-problems]

**Sub-problem Processing:** (Repeat the following steps for each
sub-problem)
- Sub-problem 1:
  * Initial approach
  * Reflection
  * Improved solution
- Sub-problem 2:
  * Initial approach
  * Reflection
  * Improved solution
- ...(adjust according to the actual number of sub-problems)

**Integration:**
[Integrate sub-problem solutions into a complete solution]

**Overall Reflection:**
[Evaluate the integrated complete solution]

**Final Optimization:**
[Make final optimizations based on overall reflection]

**Summary:**
[Summarize key points of the final comprehensive solution]

The solution section should maintain logical, accurate, and concise 
expression, detailing the steps needed to reach the conclusion, 
formatted as:
**Solution:**
[Provide the final solution here]
\end{verbatim}
\end{tcolorbox}
\section{Extra Experiment Details}
\label{appendix:experiment details}
\subsection{Training Details} 
Our training environment is equipped with 8 A800 GPUs with 80GB memory, and uses the DeepSpeed ZeRO-3 framework for distributed training. To further optimize GPU memory, we also introduced technologies such as Tensor Parallelism and liger kernel. Here, we share the YAML and DeepSpeed configuration files used for qwen2.5-Math-7B model training as an example.
\begin{tcolorbox}[title={\textbf{\small Yaml config for running Qwen2.5-Math-7B on LiteCoT}}, boxrule=2pt, arc=0mm, breakable]\begin{verbatim}
    
# Logging and Output arguments
log_level: info
logging_strategy: steps
logging_steps: 10
save_total_limit: 1
report_to:
# - tensorboard
- wandb
save_strategy: steps
save_steps: 100
output_dir: /output_path
overwrite_output_dir: true

# Model arguments
model_name_or_path: /model_path
model_revision: main
trust_remote_code: True
torch_dtype: bfloat16
attn_implementation: sdpa

# Data training arguments
dataset_name: /dataset_path
dataset_format: parquet
# dataset_configs:
# - all
max_seq_length: 8192
dataset_num_proc: 8
packing: true

# SFT trainer config
preprocessing_num_workers: 8 
seed: 233
do_train: true
num_train_epochs: 3
per_device_train_batch_size: 4
do_eval: true
eval_strategy: steps
eval_steps: 10
per_device_eval_batch_size: 1
optim: adamw_torch
learning_rate: 5.0e-5
lr_scheduler_type: cosine_with_min_lr
lr_scheduler_kwargs:
  min_lr_rate: 0.1
warmup_ratio: 0.1
weight_decay: 0.0001
gradient_accumulation_steps: 4
gradient_checkpointing: true
gradient_checkpointing_kwargs:
  use_reentrant: false
max_grad_norm: 1.0
bf16: true
use_liger: true
\end{verbatim}
\end{tcolorbox}

\begin{tcolorbox}[title={\textbf{\small DeepSpeed config for running Qwen2.5-Math-7B on LiteCoT}}, boxrule=2pt, arc=0mm, breakable]\begin{verbatim}
compute_environment: LOCAL_MACHINE
debug: false
deepspeed_config:
  deepspeed_multinode_launcher: standard
  offload_optimizer_device: none
  offload_param_device: none
  zero3_init_flag: true
  zero3_save_16bit_model: true
  zero_stage: 3
  tensor_parallel:
    tp_size: 8  
  offload_optimizer:
    device: none
    pin_memory: true
  offload_param:
    device: none
  train_batch_size: 16
  train_micro_batch_size_per_gpu: 4
  gradient_accumulation_steps: 4
  bf16:
    enabled: true
  optimizer:
    type: AdamW
    params:
      lr: 5.0e-5
      betas: [0.9, 0.999]
      eps: 1e-8
      weight_decay: 0.0001
      bias_correction: true
main_process_port: 29500
distributed_type: DEEPSPEED
downcast_bf16: 'no'
machine_rank: 0
main_training_function: main
mixed_precision: bf16
num_machines: 1
num_processes: 8
rdzv_backend: static
same_network: true
tpu_env: []
tpu_use_cluster: false
tpu_use_sudo: false
use_cpu: false
\end{verbatim}
\end{tcolorbox}

\subsection{Evaluation Details} 
In our evaluation process, the evaluation scripts mainly come from two sources. For the four benchmarks AIME24, AIME25, GPQA, and MATH500, we use Hugging Face's LightEval library \url{https://github.com/huggingface/lighteval/tree/main} for model evaluation. For the remaining benchmarks focusing on mathematical reasoning ability, including GSM8K, Minerva Math, GaoKao2023en, OlympiadBench, College Math, Sat Math, MMLU STEM, and ACM23, we use the evaluation code provided in the Qwen2.5 code repository \url{https://github.com/QwenLM/Qwen2.5-Math/tree/main/evaluation/data}.

All evaluation experiments are performed uniformly under our specific GPU environment, NVIDIA H20 with 96GB memories.

Next, we will share the shell (.sh) script files used to evaluate the AIME24, GPQA, MATH500, and AIME25 benchmarks based on Lighteval for reference.

\begin{tcolorbox}[title={\textbf{\small sh file used for evaluating AIME24, AIME25, GPQA and MATH500}}, boxrule=2pt, arc=0mm, breakable]\begin{verbatim}

MODEL=/model_path
NUM_GPUS=n
export CUDA_VISIBLE_DEVICES="0,...,n-1"
MODEL_ARGS="pretrained=$MODEL,dtype=bfloat16,tensor_parallel_size=
$NUM_GPUS,max_model_length=32768,gpu_memory_utilization=0.8
,generation_parameters={max_new_tokens:32768,temperature:0.6,top_p:0.95}"
OUTPUT_DIR=data/evals/$MODEL

# AIME 2024
TASK=aime24
lighteval vllm $MODEL_ARGS "lighteval|$TASK|0|0" \
    --use-chat-template \
    --output-dir $OUTPUT_DIR
# AIME 2025
TASK=aime25
lighteval vllm $MODEL_ARGS "lighteval|$TASK|0|0" \
    --use-chat-template \
    --output-dir $OUTPUT_DIR
# MATH-500
TASK=math_500
lighteval vllm $MODEL_ARGS "lighteval|$TASK|0|0" \
    --use-chat-template \
    --output-dir $OUTPUT_DIR
# GPQA Diamond
TASK=gpqa:diamond
lighteval vllm $MODEL_ARGS "lighteval|$TASK|0|0" \
    --use-chat-template \
    --output-dir $OUTPUT_DIR
\end{verbatim}
\end{tcolorbox}

\subsection{Benchmark Details}
In our evaluation part, we totally evaluate our models on 11 benchmarks. They are:
\begin{itemize}
    \item MATH500: Tests secondary school mathematics with open-ended problem-solving questions requiring step-by-step solutions, typically comprising 500 problems.
    \item AIME24: Assesses advanced high school mathematics (American Invitational Mathematics Examination level) through integer-answer competition-style problems, with a specific recent contest (e.g., AIME 2024) having 30 questions.
    \item GPQA: Evaluates graduate-level physics quantitative reasoning with challenging, multi-step, free-response questions written by physics experts, containing 448 questions.
    \item GSM8K: Focuses on grade school mathematics (grades 3-8) with multi-step arithmetic word problems requiring basic operations, featuring approximately 1,319 test questions.
    \item Minerva Math: Covers a broad range of quantitative reasoning from elementary mathematics to advanced levels using free-response math problems, often drawing from existing datasets and typically used with around 2,000 questions from its test set (MATH subset).
    \item GaoKao2023: Tests Chinese National College Entrance Examination (Gaokao) mathematics questions from 2023, featuring complex problems covering a wide high-school curriculum, with the math section typically having around 20-22 questions per exam paper version.
    \item OlympiadBench: Assesses mathematical Olympiad problem-solving abilities with highly challenging, proof-oriented or complex computational questions from various international and national Olympiads, containing 185 problems.
    \item College Math: Covers US college-level mathematics topics with multiple-choice and free-response questions spanning calculus, algebra, and statistics, with one common dataset having 1,760 questions.
    \item Sat Math: Evaluates high school mathematics relevant to college readiness (SAT exam) through multiple-choice and grid-in questions on algebra, problem-solving, and data analysis, with the math sections of an official SAT practice test typically containing 58 questions.
    \item MMLU STEM: A subset of the Massive Multitask Language Understanding (MMLU) benchmark, it tests expertise in STEM fields (Science, Technology, Engineering, and Mathematics) at high school and college levels via multiple-choice questions, with the STEM subset comprising several thousand questions across various specific STEM subjects (e.g., the "Mathematics" part alone has 250 dev and 1175 test questions).
    \item ACM23: Assesses competitive programming skills (similar to ACM International Collegiate Programming Contest problems) with algorithmic reasoning challenges requiring code solutions or detailed algorithmic steps, with specific contest sets varying but problem collections often containing hundreds of problems (the user might be referring to a specific year's problem set or a curated collection).
\end{itemize}

\subsection{Additional Experiment Results}

\begin{figure}[htbp]
    \centering
    \includegraphics[width=\textwidth]{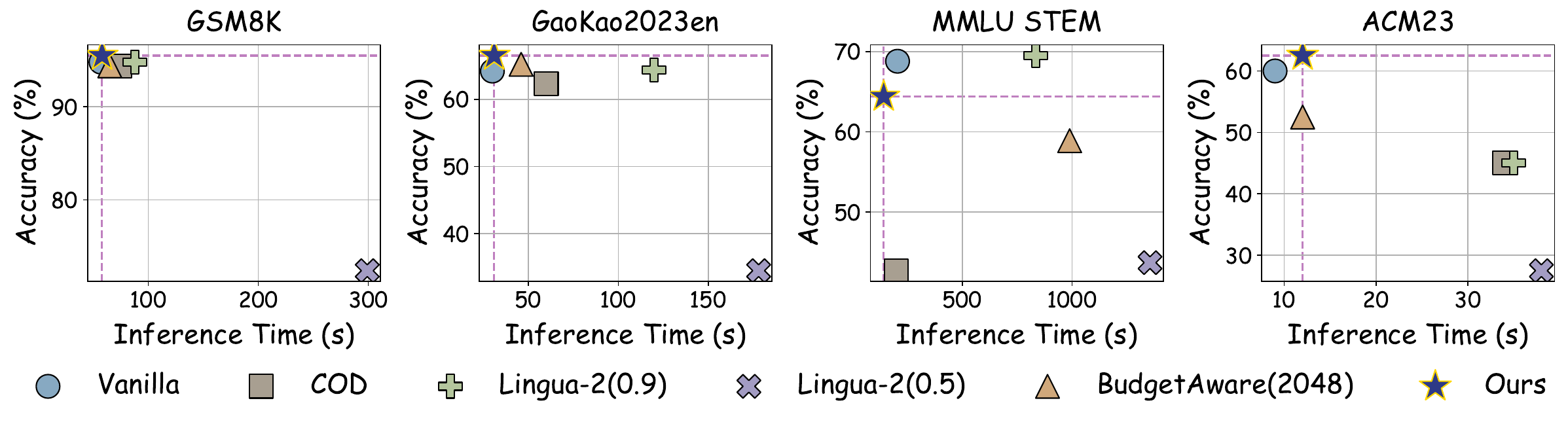}
    \vspace{-1em}
    \caption{Additional experiment results comparing inference time and accuracy of different methods across multiple benchmarks. Our method (Ours) consistently achieves optimal accuracy while significantly reducing inference time compared to existing methods.}
    \label{appendix:acc_vs_inference_time_on_long_and_short}
    \vspace{-1em}
\end{figure}

\section{Dataset Comparison}
\label{appen:dataset examples}
\subsection{Cases of Samples}
In this section, we will show the model distillation results for the same problem in different datasets. We will show the comparison of three different difficulty levels: easy, medium, and hard. Because the original answer for hard is too long, we only show the Litecot data and OpenThoughts data for easy and medium difficulty levels here.

\stitle{Easy Question:} 
\begin{verbatim}
    Return your final response within \boxed{}.
    Find all natural numbers that,
    when divided by 7, give the same quotient and remainder.

\end{verbatim}

\begin{tcolorbox}[title={\textbf{\small solutions of easy problems in LiteCoT}}, boxrule=2pt, arc=0mm, breakable]\begin{verbatim}
<|begin_of_thought|>
**Analysis:**
The problem requires finding all natural numbers \( n \) such that
when divided by 7, the quotient \( q \) equals the remainder
\( r \). Using the division algorithm, \( n = 7q + r \), 
with \( 0 \leq r < 7 \). The condition \( q = r \) must hold.

**Approach:**
Substitute \( q = r \) into the division equation to get \( n = 8r \).
Since \( r \) must be a valid remainder (1 \leq \( r \) \leq 6,
as \( r = 0 \) would make \( n = 0 \), not a natural number), compute 
\( n \) for \( r = 1, 2, ..., 6 \).

**Summary:**
The natural numbers are \( 8r \) for \( r = 1 \) to \( 6 \), yielding
\( 8, 16, 24, 32, 40, 48 \).
<|end_of_thought|>

<|begin_of_solution|>
**Solution:**
To find all natural numbers \( n \) such that when divided by 7,
the quotient equals the remainder:
1. Use the division formula: \( n = 7q + r \), where \( 0 \leq r < 7 \).
2. Set \( q = r \), leading to \( n = 7r + r = 8r \).
3. Since \( r \) must be a natural number (remainder \geq 1) and
\( r < 7 \), valid values for \( r \) are \( 1, 2, 3, 4, 5, 6 \).
4. Calculate \( n \) for each \( r \):
   - \( r = 1 \Rightarrow n = 8 \)
   - \( r = 2 \Rightarrow n = 16 \)
   - \( r = 3 \Rightarrow n = 24 \)
   - \( r = 4 \Rightarrow n = 32 \)
   - \( r = 5 \Rightarrow n = 40 \)
   - \( r = 6 \Rightarrow n = 48 \)

Thus, the natural numbers are:
\[ \boxed{8, \ 16, \ 24, \ 32, \ 40, \ 48} \]
<|end_of_solution|>

\end{verbatim}
\end{tcolorbox}

\begin{tcolorbox}[title={\textbf{\small solutions of easy problems in OpenThoughts}}, boxrule=2pt, arc=0mm, breakable]\begin{verbatim}


<|begin_of_thought|>

Okay, so I need to find all natural numbers that, when divided by 7,
give the same quotient and remainder. Let me start by understanding 
the problem properly. 

First, let's recall that when we divide a natural number by 7,
we can express it in the form of the division algorithm: n = 7q + r,
where q is the quotient and r is the remainder. The remainder r must 
satisfy 0 \len r < 7 because we're dividing by 7. So, the possible 
values for r are 0, 1, 2, 3, 4, 5, 6.

Now, the problem states that the quotient and the remainder are the same.
That means q = r. So substituting that into the equation,
we get n = 7q + q = 8q. Therefore, n must be a multiple of 8. But wait,
we need to ensure that the remainder r is indeed equal to q and that it 
satisfies the condition 0 \len r < 7. 

Since r = q, and r must be less than 7,
q must also be less than 7. Because if q were 7 or more,
then the remainder r would be 7 or more, which isn't allowed in division 
by 7. So q can be 0, 1, 2, 3, 4, 5, or 6. 

However, n is a natural number. Depending on the definition,
natural numbers usually start from 1, so n must be at least 1. If q = 0,
then n = 8*0 = 0, which might not be considered a natural number.
Let me check if 0 is included here. If the problem defines natural 
numbers as starting from 1, then q can only be from 1 to 6. But if 0 is 
included, then 0 would also be a solution. 

But let me confirm the standard definition.
In mathematics, natural numbers typically start from 1, so 0 is excluded.
Therefore, q must be at least 1. So q can be 1, 2, 3, 4, 5, 6. Therefore, 
the corresponding n values would be 8*1=8, 8*2=16, 8*3=24, 8*4=32, 
8*5=40, 8*6=48. 

But wait, I need to verify if each of these numbers actually satisfies 
the original condition. Let's take each n and divide by 7 to see if the
quotient and remainder are equal. 

Starting with n=8: 8 divided by 7 is 1 with a remainder of 1.
Here, quotient q=1 and remainder r=1, so they are equal. That works. 

Next, n=16: 16 divided by 7 is 2 with a remainder of 2. So q=2 and r=2.
That's good. 

n=24: 24 divided by 7 is 3 with a remainder of 3. Because 7*3=21,
24-21=3. So q=3, r=3. Correct. 

n=32: 32 divided by 7 is 4 with remainder 4 (7*4=28, 32-28=4). 
So q=4, r=4. 

n=40: 40 divided by 7 is 5 with remainder 5. 7*5=35, 40-35=5. 
So q=5, r=5. 

n=48: 48 divided by 7 is 6 with remainder 6. 7*6=42, 48-42=6. 
So q=6, r=6. 

Okay, all of these check out. Now, if we consider q=7, then n=8*7=56.
Let's divide 56 by 7: quotient is 8, remainder is 0. Here, q would have 
been 8 (but according to our original substitution, q=r=7). Wait, 
this is where the problem arises. If q=7, then r=7, but 7 is not a valid 
remainder when dividing by 7. The remainder must be less than 7.
Therefore, q=7 is invalid because it would lead to a remainder of 7, 
which isn't allowed. Hence, the maximum q can be is 6. 

Therefore, the natural numbers are 8, 16, 24, 32, 40, 48. 

Wait a second, let me make sure there are no other numbers beyond these 
that could satisfy the condition. Suppose someone claims that there is a 
larger number where q=r. Let's say q=7, but as I saw, r would have to be 
7, which is not allowed. Similarly, for q=8, r=8 which is even more 
invalid. So the only possible q values are 1 through 6,
leading to n=8q where q=1 to 6. 

But just to be thorough, let's check if there's a different approach. 
Suppose we set up the equation:

When n is divided by 7, quotient q equals remainder r. 
So, by division algorithm:

n = 7q + r

But since q = r, substitute:

n = 7q + q = 8q

Additionally, the remainder r must satisfy 0 \len r < 7. Since r = q,
this translates to 0 \len q < 7. However, since n is a natural number
(assuming starting at 1), q must be at least 1. Therefore, q
{1,2,3,4,5,6}, leading to n  {8,16,24,32,40,48}.

Therefore, these are all the natural numbers that satisfy the condition. 
Let me check if n=0 is allowed. If the problem includes 0 as a natural
number, which is sometimes the case in certain contexts, then 0 would be
a solution as well (0 divided by 7 is 0 with remainder 0). However, 
since the problem specifies "natural numbers," and given that in most 
mathematical contexts natural numbers start from 1, we can exclude 0.

Thus, the final answer should be the numbers 8, 16, 24, 32, 40, and 48.

<|end_of_thought|>

<|begin_of_solution|>

To find all natural numbers \( n \) that, when divided by 7, yield the
same quotient \( q \) and remainder \( r \), we start with the division 
algorithm:  
\[ n = 7q + r \]  
where \( 0 \leq r < 7 \). The problem states \( q = r \),
so substituting \( r = q \) gives:  
\[ n = 7q + q = 8q. \]  

Since \( r = q \) must satisfy \( 0 \leq q < 7 \), 
and \( n \) is a natural number (starting from 1),
\( q \) must be in \( \{1, 2, 3, 4, 5, 6\} \).
Substituting these values into \( n = 8q \):  

- For \( q = 1 \): \( n = 8 \times 1 = 8 \).  
- For \( q = 2 \): \( n = 8 \times 2 = 16 \).  
- For \( q = 3 \): \( n = 8 \times 3 = 24 \).  
- For \( q = 4 \): \( n = 8 \times 4 = 32 \).  
- For \( q = 5 \): \( n = 8 \times 5 = 40 \).  
- For \( q = 6 \): \( n = 8 \times 6 = 48 \).  

Verification confirms each \( n \) satisfies \( \text{quotient} = 
\text{remainder} \) when divided by 7. For example:
- \( 8 \div 7 = 1 \) with remainder \( 1 \),  
- \( 16 \div 7 = 2 \) with remainder \( 2 \), etc.  

**Final Answer:**  
The natural numbers are \(\boxed{8, 16, 24, 32, 40, 48}\).

<|end_of_solution|>
\end{verbatim}
\end{tcolorbox}

\stitle{Medium Question:} 
\begin{verbatim}
    Return your final response within \boxed{}.
    Let \( h_{1}, h_{2}, h_{3} \) be the heights of a triangle,
    and \( r \) be the radius of the inscribed circle.
    Prove that \( h_{1} + h_{2} + h_{3} \geq 9r \).
\end{verbatim}

\begin{tcolorbox}[title={\textbf{\small solutions of medium problems in LiteCoT}}, boxrule=2pt, arc=0mm, breakable]\begin{verbatim}
<|begin_of_thought|>
**Analysis:**
The problem requires proving an inequality involving triangle heights
and inradius. Key elements are the relationships between heights (h_1, 
h_2, h_3), inradius (r), area (S), and semiperimeter (p). The challenge 
lies in connecting these quantities through known formulas and applying 
an inequality.

**Initial Approach:**
Express heights in terms of area and sides (h_i = 2S/a_i). Use the 
identity S = p·r to substitute into the sum of heights. This leads to 
h_1 + h_2 + h_3 = 2pr(1/a + 1/b + 1/c). To establish the inequality, 
relate 1/a + 1/b + 1/c to semiperimeter via Cauchy-Schwarz.

**Reflection:**
The Cauchy-Schwarz inequality (for vectors (a, b, c) and (1/a, 1/b, 1/c))
provides (a + b + c)(1/a + 1/b + 1/c) \geq 9. Since a + b + c = 2p, this 
bounds 1/a + 1/b + 1/c \geq 9/(2p). Substituting back confirms 
the required inequality.

**Improvement:**
Verify the validity of Cauchy-Schwarz application and ensure all
substitutions align with triangle properties. No logical gaps detected.

**Summary:**
The proof hinges on expressing heights via area, applying Cauchy-Schwarz 
to relate reciprocals of sides to semiperimeter, and substituting into
the sum of heights to derive the inequality.
<|end_of_thought|>

<|begin_of_solution|>
**Solution:**
1. **Express heights via area**: For a triangle with sides \(a, b, c\) 
and corresponding heights \(h_1, h_2, h_3\), the area \(S\) is:
   \[
   h_1 = \frac{2S}{a},\quad h_2 = \frac{2S}{b},\quad h_3 = \frac{2S}{c}.
   \]

2. **Sum of heights**:
   \[
   h_1 + h_2 + h_3 = 2S\left(\frac{1}{a} + \frac{1}{b} + 
   \frac{1}{c}\right).
   \]

3. **Relate \(S\) to inradius \(r\)**:
Using \(S = p \cdot r\) (where \(p = \frac{a+b+c}{2}\) is semiperimeter):
   \[
   h_1 + h_2 + h_3 = 2pr\left(\frac{1}{a} + \frac{1}{b} + 
   \frac{1}{c}\right).
   \]

4. **Apply Cauchy-Schwarz inequality**:
   \[
   (a + b + c)\left(\frac{1}{a} + \frac{1}{b} + \frac{1}{c}\right) 
   \geq (1 + 1 + 1)^2 = 9.
   \]
   Since \(a + b + c = 2p\), this simplifies to:
   \[
   \frac{1}{a} + \frac{1}{b} + \frac{1}{c} \geq \frac{9}{2p}.
   \]

5. **Substitute into height sum**:
   \[
   h_1 + h_2 + h_3 \geq 2pr \cdot \frac{9}{2p} = 9r.
   \]

Thus, the inequality \(h_1 + h_2 + h_3 \geq 9r\) is proven.

\[
\boxed{h_{1} + h_{2} + h_{3} \geq 9r}
\]
<|end_of_solution|>
\end{verbatim}
\end{tcolorbox}

\begin{tcolorbox}[title={\textbf{\small solutions of medium problems in OpenThoughts}}, boxrule=2pt, arc=0mm, breakable]\begin{verbatim}
<|begin_of_thought|>

Alright, so I need to prove that the sum of the heights of a
triangle 
is at least nine times the inradius. Let me start by recalling some 
formulas related to triangles, heights, and the inradius. 

First, the height (altitude) corresponding to a side of a triangle
is given by the formula \( h_a = \frac{2\Delta}{a} \), where
\( \Delta \) is the area of the triangle and \( a \) is the length
of 
the side. Similarly, \( h_b = \frac{2\Delta}{b} \) and 
\( h_c = \frac{2\Delta}{c} \). So, the sum of the heights
\( h_1 + h_2 + h_3 \) would be \( \frac{2\Delta}{a} +
\frac{2\Delta}{b
} 
+ \frac{2\Delta}{c} \). 

On the other hand, the inradius \( r \) is related to the area and
the
semiperimeter \( s \) by the formula \( r = \frac{\Delta}{s} \),
where
 
\( s = \frac{a + b + c}{2} \). So, the inradius is the area divided
by
 
the semiperimeter.

Our goal is to show that \( h_1 + h_2 + h_3 \geq 9r \). Let's
substitu
te
the expressions for the heights and the inradius in terms of the
area 
and the sides. 

First, express \( h_1 + h_2 + h_3 \) as \( 2\Delta \left(
\frac{1}{a} 
+
\frac{1}{b} + \frac{1}{c} \right) \). The inradius \( r \) is
\( \frac{\Delta}{s} \), so \( 9r = \frac{9\Delta}{s} \). 
Therefore, the inequality we need to prove becomes:

\( 2\Delta \left( \frac{1}{a} + \frac{1}{b} + \frac{1}{c} \right)
\geq
\frac{9\Delta}{s} \)

Since \( \Delta \) is positive (it's the area of a triangle), we can
divide both sides by \( \Delta \) without changing the inequality 
direction:

\( 2 \left( \frac{1}{a} + \frac{1}{b} + \frac{1}{c} \right) \geq
\frac{9}{s} \)

Simplifying further, multiply both sides by \( s \):

\( 2s \left( \frac{1}{a} + \frac{1}{b} + \frac{1}{c} \right) \geq 9
\)

So, the inequality reduces to proving that \( 2s \left( \frac{1}{a}
+
\frac{1}{b} + \frac{1}{c} \right) \geq 9 \), where 
\( s = \frac{a + b + c}{2} \).

Let me write this more explicitly. Substitute \( s = \frac{a + b +
c}{
2}
\), so:

\( 2 \times \frac{a + b + c}{2} \times \left( \frac{1}{a} +
\frac{1}{b
}
+ \frac{1}{c} \right) \geq 9 \)

The 2 and the denominator 2 cancel out, so:

\( (a + b + c) \left( \frac{1}{a} + \frac{1}{b} + \frac{1}{c}
\right) 
\geq 9 \)

Ah, now this looks familiar. This is a classic inequality that might
b
e 
related to the AM-HM inequality or Cauchy-Schwarz. 
Let me recall the Cauchy-Schwarz inequality in the form:

\( (x_1y_1 + x_2y_2 + x_3y_3)^2 \leq (x_1^2 + x_2^2 + x_3^2)(y_1^2 +
y
_2^2 + y_3
^2) \)

But perhaps another form is more useful here. The Cauchy-Schwarz 
inequality for sequences \( (a, b, c) \) and \( \left( \frac{1}{a},
\frac{1}{b}, \frac{1}{c} \right) \) gives:

\( \left( a \times \frac{1}{a} + b \times \frac{1}{b} + c \times
\frac
{1}{c} \ri
ght)^2 \leq (a^2 + b^2 + c^2) \left( \frac{1}{a^2} + \frac{1}{b^2} +
\
frac{1}{c^
2} \right) \)

But this simplifies to \( (1 + 1 + 1)^2 \leq (a^2 + b^2 + c^2)
\left( 
\frac{1}{a
^2} + \frac{1}{b^2} + \frac{1}{c^2} \right) \), which is 9 \leq
someth
ing. Not s
ure if that's directly helpful here.

Alternatively, consider the AM-HM inequality for three positive real
n
umbers:

\( \frac{a + b + c}{3} \geq \frac{3}{\frac{1}{a} + \frac{1}{b} +
\frac
{1}{c}} \)

Rearranging this gives:

\( (a + b + c)\left( \frac{1}{a} + \frac{1}{b} + \frac{1}{c} \right)
\
geq 9 \)

Which is exactly the inequality we need! So, according to the AM-HM
in
equality, 
the product of the sum of the sides and the sum of their reciprocals
i
s indeed g
reater than or equal to 9. Therefore, the original inequality \( h_1
+
 h_2 + h_3
 \geq 9r \) holds true.

But wait, let me double-check the steps to make sure I didn't skip
any
thing cruc
ial. Starting from the heights:

\( h_1 = \frac{2\Delta}{a} \), similarly for h2, h3. Then sum is \(
2\
Delta(\fra
c{1}{a} + \frac{1}{b} + \frac{1}{c}) \). Then, r is \(
\frac{\Delta}{s
} \), so 9
r is \( \frac{9\Delta}{s} \). Therefore, the inequality becomes:

\( 2\Delta(\frac{1}{a} + \frac{1}{b} + \frac{1}{c}) \geq
\frac{9\Delta
}{s} \)

Cancel \Delta:

\( 2(\frac{1}{a} + \frac{1}{b} + \frac{1}{c}) \geq \frac{9}{s} \)

Multiply both sides by s:

\( 2s(\frac{1}{a} + \frac{1}{b} + \frac{1}{c}) \geq 9 \)

But since s = (a + b + c)/2, substituting:

\( 2*(a + b + c)/2 * (sum reciprocals) = (a + b + c)(sum
reciprocals) 
\geq 9 \)

Which is exactly the AM-HM inequality. So, yes, the steps are
correct.

But hold on, AM-HM gives (a + b + c)/3 \geq 3/(1/a + 1/b + 1/c),
which
 rearrange
s to (a + b + c)(1/a + 1/b + 1/c) \geq 9. So that's exactly the
inequa
lity we ha
ve here. Therefore, equality holds when a = b = c, i.e., in an
equilat
eral trian
gle. So, this inequality is tight for equilateral triangles.

Therefore, the original inequality \( h_1 + h_2 + h_3 \geq 9r \) is
pr
oven by ap
plying the AM-HM inequality to the sides of the triangle. 

Alternatively, another approach might involve expressing the heights
i
n terms of
 the inradius and other elements, but since this path worked out,
mayb
e that's s
ufficient. However, to be thorough, let me consider if there are
other
 ways or p
ossible pitfalls.

Wait, another thought: The formula for the inradius is r = \Delta/s,
s
o s = \Del
ta/r. Then, since h1 + h2 + h3 = 2\Delta(1/a + 1/b + 1/c), we can
writ
e the ineq
uality as:

2\Delta(1/a + 1/b + 1/c) \geq 9r = 9\Delta/s

Cancelling \Delta, we get 2(1/a + 1/b + 1/c) \geq 9/s, which is
equiva
lent to 2s
(1/a + 1/b + 1/c) \geq 9. Then substituting s = (a + b + c)/2, we
have
:

2*(a + b + c)/2*(1/a + 1/b + 1/c) = (a + b + c)(1/a + 1/b + 1/c)
\geq 
9

Which again is the same inequality. So, regardless of the path, we
end
 up with t
he AM-HM inequality.

Alternatively, if I didn't recall AM-HM, maybe using Cauchy-Schwarz:

( (sqrt(a))^2 + (sqrt(b))^2 + (sqrt(c))^2 ) * ( (1/sqrt(a))^2 +
(1/sqr
t(b))^2 + 
(1/sqrt(c))^2 ) \geq (1 + 1 + 1)^2 = 9

But that's the same as (a + b + c)(1/a + 1/b + 1/c) \geq 9. So,
Cauchy
-Schwarz a
lso gives the same result. Therefore, the inequality holds by
Cauchy-S
chwarz as 
well.

Therefore, regardless of the method used, the key inequality needed
is
 (a + b + 
c)(1/a + 1/b + 1/c) \geq 9, which is a standard result. Therefore,
the
 original 
inequality is proven.

Another angle: Let's consider specific cases. For an equilateral
trian
gle, let's
 check if equality holds. Suppose all sides are equal, say length a.
T
hen, the s
emiperimeter s = 3a/2. The inradius r = \Delta/s. The area \Delta =
(s
qrt(3)/4)a
^2. So r = (sqrt(3)/4 a^2)/(3a/2) ) = (sqrt(3)/4 a^2) * (2)/(3a) ) =
(
sqrt(3)/6)
a. The heights h1, h2, h3 are all equal to (sqrt(3)/2)a. So sum of
hei
ghts is 3*
(sqrt(3)/2)a = (3sqrt(3)/2)a. Then 9r = 9*(sqrt(3)/6)a =
(3sqrt(3)/2)a
. Therefor
e, sum of heights equals 9r in this case, so equality holds, which
mat
ches the c
ondition when AM-HM equality holds, i.e., when all sides are equal.

For a non-equilateral triangle, let's take a simple case. Suppose a
= 
3, b = 4, 
c = 5 (right-angled triangle). Then, semiperimeter s = (3 + 4 + 5)/2
=
 6. Area \
Delta = (3*4)/2 = 6. Inradius r = \Delta/s = 6/6 = 1. Heights: h1 =
2\
Delta/a = 
12/3 = 4, h2 = 12/4 = 3, h3 = 12/5 = 2.4. Sum of heights: 4 + 3 +
2.4 
= 9.4. 9r 
= 9*1 = 9. So 9.4 \geq 9, which holds. So the inequality is
satisfied 
here.

Another example: a = 2, b = 2, c = 3. Then s = (2 + 2 + 3)/2 = 3.5.
Ar
ea using H
eron's formula: sqrt(3.5*(3.5 - 2)*(3.5 - 2)*(3.5 - 3)) =
sqrt(3.5*1.5
*1.5*0.5) 
= sqrt(3.5*1.5*0.75) = sqrt( (3.5*0.75)*1.5 ) = sqrt(2.625*1.5)
\appro
x sqrt(3.9
375) \approx 1.984. So r = 1.984 / 3.5 \approx 0.567. Heights: h1 =
2\
Delta/a \a
pprox 2*1.984 /2 \approx 1.984, h2 same as h1, h3 = 2*1.984 /3
\approx
 1.323. Su
m \approx 1.984 + 1.984 + 1.323 \approx 5.291. 9r \approx 5.103. So
5.
291 \geq 5
.103. Holds as well.

Wait, but according to the calculation, the sum of heights is
approxim
ately 5.29
1, which is greater than 9r \approx 5.103. So, even in this case,
the 
inequality
 holds. 

But let me check a more extreme triangle. Suppose a triangle with
side
s approach
ing degeneracy. Let’s take a = 1, b = 1, c approaching 2 (but still
a 
valid tria
ngle). Let’s compute when c approaches 2, say c = 2 - \epsilon,
where 
\epsilon i
s very small. Then semiperimeter s = (1 + 1 + 2 - \epsilon)/2 = (4 -
\
epsilon)/2
 = 2 - \epsilon/2. Area \Delta using Heron's formula: sqrt( s(s -
a)(s
 - b)(s - 
c) ) = sqrt( (2 - \epsilon/2)(2 - \epsilon/2 -1)(2 - \epsilon/2
-1)(2 
- \epsilon
/2 - (2 - \epsilon)) ) = sqrt( (2 - \epsilon/2)(1 - \epsilon/2)(1 -
\e
psilon/2)(
\epsilon/2) ). As \epsilon approaches 0, this becomes sqrt(2*1*1*0)
= 
0. So \Del
ta approaches 0, and r = \Delta/s approaches 0. The heights: h1 and
h2
 (correspo
nding to sides a=1 and b=1) will be h1 = 2\Delta/a = 2\Delta/1 =
2\Del
ta, h2 sim
ilarly 2\Delta, and h3 = 2\Delta/c \approx  2\Delta/(2 - \epsilon)
\ap
prox  \Del
ta. So sum of heights \approx  2\Delta + 2\Delta + \Delta = 5\Delta.
9
r \approx 
 9*(\Delta/s) \approx  9*(\Delta/(2 - \epsilon/2)) \approx
(9/2)\Delt
a. So 5\De
lta vs (9/2)\Delta. Since 5 > 4.5, then 5\Delta \geq (9/2)\Delta
when 
\Delta > 0
. But as \Delta approaches 0, both sides approach 0, but 5\Delta is
st
ill greate
r. However, in this limit, the inequality holds because 5\Delta \geq
(
9/2)\Delta
 to 5 \geq 4.5, which is true. But this is a bit hand-wavy because
whe
n \epsilon
 approaches 0, the triangle becomes degenerate. However, for a valid
n
on-degener
ate triangle, \epsilon must be greater than 0, so \Delta is
positive, 
and the in
equality holds.

Alternatively, consider a very skewed triangle where one side is
much 
longer tha
n the others. But perhaps in all cases, the sum of the heights is
stil
l greater 
than 9r. 

Wait, but in the example with sides 2,2,3, the sum of heights was
appr
oximately 
5.291 and 9r \approx 5.103. The difference isn't huge, but the
inequal
ity holds.
 So even in a relatively "flat" triangle, the inequality still
holds. 

Therefore, it seems that through multiple examples and applying the
AM
-HM inequa
lity, the original statement is valid. 

Another approach could be using trigonometric identities. Let’s
denote
 the trian
gle with sides a, b, c opposite to angles A, B, C respectively. The
he
ights can 
be expressed as h_a = b sin C = c sin B, similarly for others. But
thi
s might co
mplicate things. Alternatively, since r = (a + b - c)/2 * tan(C/2)
etc
., but per
haps this is not useful here.

Alternatively, express everything in terms of r and the
semiperimeter.
 Since r =
 \Delta/s, and \Delta = r*s. Then, the heights can be written as h_a
=
 2\Delta/a
 = 2r*s/a. Similarly, h_b = 2r*s/b, h_c = 2r*s/c. Therefore, sum of
he
ights is 2
r*s (1/a + 1/b + 1/c). We need to show that this is \geq9r. Divide
bot
h sides by
 r (positive), so need to show 2s(1/a + 1/b + 1/c) \geq9, which is
the
 same as b
efore. So again, we come back to the same inequality. 

Therefore, regardless of the approach, the crux is proving (a + b +
c)
(1/a + 1/b
 + 1/c) \geq9, which is a direct consequence of the AM-HM
inequality. 
Hence, the
 original inequality holds.

In conclusion, the key steps are:

1. Express the heights in terms of the area and sides.
2. Express the inradius in terms of the area and semiperimeter.
3. Formulate the inequality in terms of the sides and semiperimeter.
4. Apply the AM-HM inequality to establish the required result.
5. Verify with specific examples to ensure validity.

Therefore, the inequality \( h_1 + h_2 + h_3 \geq 9r \) is proven.

<|end_of_thought|>

<|begin_of_solution|>

To prove that the sum of the heights of a triangle \( h_1 + h_2 +
h_3 
\) is at l
east nine times the inradius \( r \), we proceed as follows:

1. **Express Heights and Inradius**:
   - Heights: \( h_a = \frac{2\Delta}{a}, \, h_b =
\frac{2\Delta}{b}, 
\, h_c = \
frac{2\Delta}{c} \), where \( \Delta \) is the area and \( a, b, c
\) 
are the si
des.
   - Sum of heights: \( h_1 + h_2 + h_3 = 2\Delta\left(\frac{1}{a} +
\
frac{1}{b}
 + \frac{1}{c}\right) \).
   - Inradius: \( r = \frac{\Delta}{s} \), where \( s = \frac{a + b
+ 
c}{2} \) i
s the semiperimeter.

2. **Formulate the Inequality**:
   Substitute \( h_1 + h_2 + h_3 \) and \( 9r \) into the
inequality:
   \[
   2\Delta\left(\frac{1}{a} + \frac{1}{b} + \frac{1}{c}\right) \geq
\f
rac{9\Delt
a}{s}.
   \]
   Cancel \( \Delta \) and multiply through by \( s \):
   \[
   2s\left(\frac{1}{a} + \frac{1}{b} + \frac{1}{c}\right) \geq 9.
   \]

3. **Simplify Using Semiperimeter**:
   Substitute \( s = \frac{a + b + c}{2} \):
   \[
   (a + b + c)\left(\frac{1}{a} + \frac{1}{b} + \frac{1}{c}\right)
\ge
q 9.
   \]

4. **Apply the AM-HM Inequality**:
   The Arithmetic Mean-Harmonic Mean (AM-HM) inequality states:
   \[
   \frac{a + b + c}{3} \geq \frac{3}{\frac{1}{a} + \frac{1}{b} +
\frac
{1}{c}},
   \]
   which rearranges to:
   \[
   (a + b + c)\left(\frac{1}{a} + \frac{1}{b} + \frac{1}{c}\right)
\ge
q 9.
   \]
   Equality holds if and only if \( a = b = c \) (equilateral
triangle
).

5. **Conclusion**:
   The inequality \( h_1 + h_2 + h_3 \geq 9r \) follows directly
from 
the AM-HM 
inequality applied to the sides of the triangle. Examples and edge
cas
es confirm
 the validity of the result.

**Final Answer**:
\boxed{h_{1} + h_{2} + h_{3} \geq 9r}

<|end_of_solution|>
\end{verbatim}
\end{tcolorbox}

\newpage

\end{document}